\documentclass[journal]{IEEEtran}
\usepackage{times}
\usepackage{epsfig}
\usepackage{graphicx}
\usepackage{amsmath}
\usepackage{amssymb}
\usepackage{multirow}
\usepackage{color}
\usepackage[table]{xcolor}
\usepackage{overpic}
\usepackage{wrapfig,lipsum,booktabs}
\usepackage{url}
\usepackage[nocompress]{cite} 
\usepackage{tabularx}
\usepackage{ragged2e}
\usepackage{pifont}
\usepackage{threeparttable}

\newcommand{\tabFormat}{\centering \renewcommand{\arraystretch}{1.2}}

\ifdefined \GramaCheck
\newcommand{\CheckRmv}[1]{}
\newcommand{\figref}[1]{Figure 1}%
\newcommand{\tabref}[1]{Table 1}%
\newcommand{\secref}[1]{Section 1}
\renewcommand{\eqref}[1]{Equation 1}
\else
\newcommand{\CheckRmv}[1]{#1}
\newcommand{\figref}[1]{Fig.~\ref{#1}}%
\newcommand{\tabref}[1]{Table~\ref{#1}}%
\newcommand{\secref}[1]{Sec.~\ref{#1}}
\renewcommand{\eqref}[1]{Equation~(\ref{#1})}
\newcommand{\cbname}{CB-}

\usepackage[pagebackref=true,breaklinks=true,colorlinks,bookmarks=false]{hyperref}

\usepackage{xspace}
\usepackage{caption}
\usepackage{subcaption}

\newcommand{\ttl}[1]{{\color{black}{#1}}}
\newcommand{\ttlv}[1]{{\color{black}{#1}}}

\makeatletter
\DeclareRobustCommand\onedot{\futurelet\@let@token\@onedot}
\def\@onedot{\ifx\@let@token.\else.\null\fi\xspace}

\def\eg{\emph{e.g}\onedot} 
\def\ie{\emph{i.e}\onedot}

\makeatother

\fi
\ifCLASSOPTIONcompsoc
  \usepackage[nocompress]{cite}
\else
  \usepackage{cite}
\fi
\ifCLASSINFOpdf
\else
\fi
\begin{document}
%
\title{CBNet: A Composite Backbone Network Architecture for Object Detection}
%
%
%
%

\author{
        Tingting~Liang$^\ast$,~Xiaojie~Chu$^\ast$,~Yudong~Liu$^\ast$,~Yongtao~Wang,~Zhi~Tang,~Wei~Chu,~Jingdong~Chen,~Haibin~Ling
\IEEEcompsocitemizethanks{\IEEEcompsocthanksitem 
Corresponding author: Yongtao~Wang. $^\ast$ indicates equal contribution.
\IEEEcompsocthanksitem 
T. Liang, X. Chu, Y. Liu, Y. Wang, Z. Tang are with the Wangxuan Institute of Computer Technology, Peking University, Beijing 100080, China. E-mail: \{tingtingliang, bahuangliuhe, wyt, tangzhi\}@pku.edu.cn, chuxiaojie@stu.pku.edu.cn
\IEEEcompsocthanksitem 
W. Chu, J. Cheng are with Ant Group of Alibaba, China. E-mail: weichu.cw@alibaba-inc.com, jingdongchen.cjd@antfin.com
\IEEEcompsocthanksitem 
H. Ling is with the Department of Computer Science, Stony Brook University, Stony Brook, NY 11794, USA. E-mail: hling@cs.stonybrook.edu.
\IEEEcompsocthanksitem A preliminary version of this manuscript was published in \cite{LiuWWLZTL20}.
}

}

\IEEEtitleabstractindextext{%
\begin{abstract}
Modern top-performing object detectors depend heavily on backbone networks, whose advances bring consistent performance gains through exploring more effective network structures.  
In this paper, we propose a novel and flexible backbone framework, namely \textit{CBNet}, to construct high-performance detectors using \textit{existing} open-source pre-trained backbones under the pre-training fine-tuning paradigm.  In particular, CBNet architecture groups multiple identical backbones, which are connected through composite connections. Specifically, it integrates the high- and low-level features of multiple identical backbone networks and gradually expands the receptive field to more effectively perform object detection. We also propose a better training strategy with \textit{auxiliary supervision} for CBNet-based detectors. CBNet has strong generalization capabilities for different backbones and head designs of the detector architecture.
Without additional pre-training of the composite backbone, CBNet can be adapted to various backbones (\textit{i.e.,} CNN-based vs. Transformer-based) and head designs of most mainstream detectors (\textit{i.e.,} one-stage vs. two-stage, anchor-based vs. anchor-free-based). Experiments provide strong evidence that, compared with simply increasing the depth and width of the network, CBNet introduces a more efficient, effective, and resource-friendly way to build high-performance backbone networks.
Particularly, our \cbname Swin-L achieves 59.4\% box AP and 51.6\% mask AP on COCO \texttt{test-dev} under the single-model and single-scale testing protocol, which are significantly better than the state-of-the-art results (\textit{i.e.,} 57.7\% box AP and 50.2\% mask AP) achieved by Swin-L, while reducing the training time by 6$\times$. With multi-scale testing, we push the current best single model result to a new record of 60.1\% box AP and 52.3\% mask AP without using extra training data. Code is available at \url{https://github.com/VDIGPKU/CBNetV2}.
\end{abstract}

}

\maketitle

\IEEEdisplaynontitleabstractindextext

%
\IEEEpeerreviewmaketitle

\CheckRmv{
	\section{Introduction}\label{sec:introduction}}
\IEEEPARstart{O}{bject} detection aims to locate each object instance from a predefined set of classes in an arbitrary image. It serves a wide range of applications such as autonomous driving, intelligent video surveillance, remote sensing, \textit{etc}. 
In recent years, great progresses have been made for object detection thanks to the booming development of deep convolutional networks \cite{krizhevsky2012imagenet}, and  excellent detectors have been proposed, \textit{e.g.}, SSD  \cite{LiuAESRFB16}, YOLO \cite{Redmon_2016_CVPR}, Faster R-CNN \cite{RenHGS15}, RetinaNet\cite{LinGGHD17}, ATSS \cite{ZhangCYLL20}, Mask R-CNN \cite{HeGDG17}, Cascade R-CNN \cite{CaiV18}, \textit{etc}. 

\CheckRmv{
\begin{figure*}[tbp]
\centering
\includegraphics[width=\textwidth]{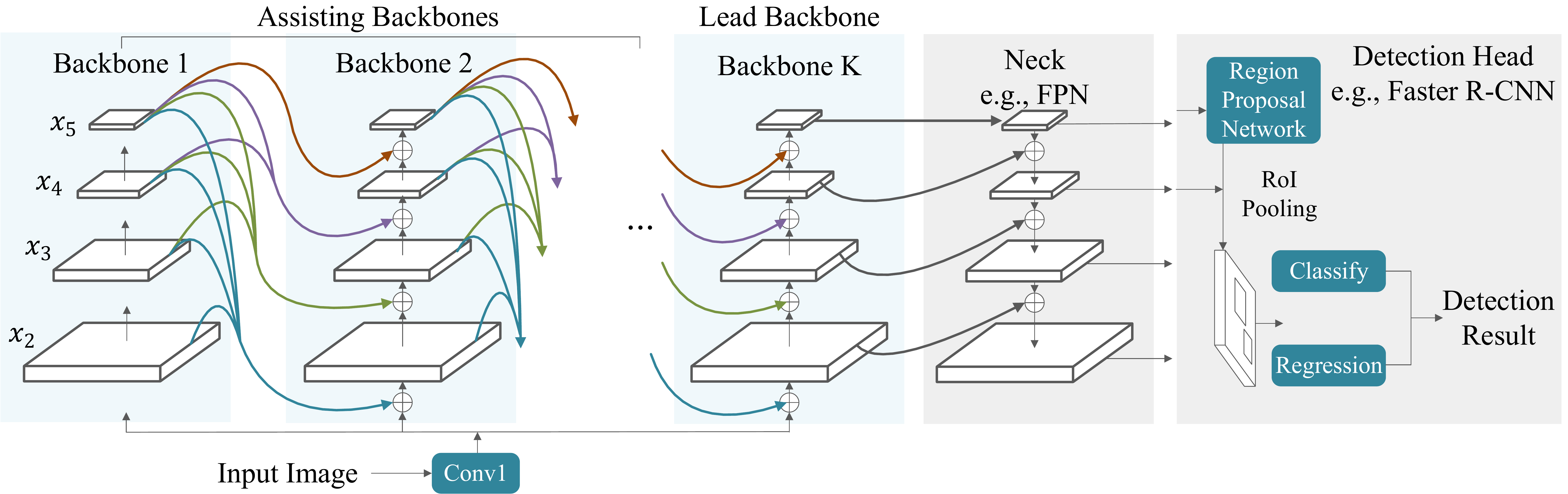}
\caption{Illustration of the proposed Composite Backbone Network  (CBNet) architecture for object detection.}
\label{fig:shouye}
\end{figure*}
}
Typically, in a Neural Network (NN)-based detector, a backbone network is used to extract basic features for detecting objects, and in most cases designed originally for image classification and pre-trained on ImageNet \cite{deng2009imagenet}. Intuitively, the more representative features extracted by the backbone, the better the performance of its host detector. 
To obtain higher accuracy, deeper and wider backbones have been exploited by mainstream detectors (\ie, from mobile-size models \cite{HowardZCKWWAA17, ZhangZLS18} and ResNet \cite{HeZRS16}, to ResNeXt \cite{xie2017aggregated} and Res2Net \cite{GaoCZZYT21}).
Recently, Transformer \cite{VaswaniSPUJGKP17,WangX0FSLL0021} based backbones have shown very promising performance. Overall, advances in large backbone pre-training demonstrate a trend towards more effective multi-scale representations in object detection.

Encouraged by the results achieved by pre-trained large backbone-based detectors, we seek further improvement to construct high-performance detectors by exploiting existing well-designed backbone architectures and their pre-trained weights. 
Though one may design a new improved backbone, the expertise and computing resources overhead can be expensive. On the one hand, designing a new architecture of backbone requires expert experience and a lot of trials and errors. On the other hand,  pre-training a new backbone (especially for large models) on ImageNet requires a large number of computational resources, which makes it costly to obtain better detection performance following the pre-training and fine-tuning paradigm.
Alternatively, training detectors from scratch saves the cost of pre-training but requires even more computing resources and training skills \cite{abs-2106-03112}.  

In this paper, we present a simple and novel composition approach to use existing pre-trained backbones under the pre-training fine-tuning paradigm. Unlike most previous methods that focus on modular crafting and require pre-training on ImageNet to strengthen the representation, we improve the existing backbone representation ability without additional pre-training.
As shown in \figref{fig:shouye}, our solution, named \textit{Composite Backbone Network} (CBNet), groups multiple identical backbones together. Specifically, parallel backbones (named assisting backbones and lead backbone) are connected via \textit{composite connections}. From left to right in \figref{fig:shouye}, the output of each stage in an assisting backbone flows to the parallel and lower-level stages of its succeeding sibling. Finally, the features of the lead backbone are fed to the neck and detection head for bounding box regression and classification.  Contrary to simple network deepening or widening, CBNet integrates the high- and low-level features of multiple backbone networks and progressively expands the receptive field for more effective object detection. Notably, each composed backbone of CBNet is \textit{initialized by the weights of an existing open-source pre-trained individual backbone}  (\eg, \cbname ResNet50 is initialized by the weights of ResNet50~\cite{HeZRS16}, which are available in the open-source community).
In addition, to further exploit the potential of CBNet, we propose an effective training strategy with supervision for assisting backbones, achieving higher detection accuracy while sacrificing no inference speed. In particular, we propose a pruning strategy to reduce the model complexity while not sacrificing accuracy.

We present two versions of CBNet. The first, named CBNetV1 \cite{LiuWWLZTL20}, connects only the adjacent stages of parallel backbones, providing a simple implementation of our composite backbone that is easy to follow. The other one, CBNetV2, combines the dense higher-level composition strategy, the auxiliary supervision, and a special pruning strategy, to fully explore the potential of CBNet for object detection. We empirically demonstrate the superiority of CBNetV2 over CBNetV1.

We demonstrate the effectiveness of our framework by conducting experiments on the challenging MS COCO benchmark~\cite{LinMBHPRDZ14}. Experiments show that CBNet has strong generalization capabilities for different backbones and head designs of the detector architecture, which enables us to train detectors that significantly outperform detectors based on larger backbones. 
Specifically, CBNet can be applied to various backbones, from convolution-based~\cite{HeZRS16, xie2017aggregated, GaoCZZYT21} to Transformer-based~\cite{abs-2103-14030}. Compared to the original backbones, CBNet boosts their performances by 3.4\%$\sim$3.5\% AP, demonstrating the effectiveness of the proposed CBNet. At comparable model complexity, our CBNet still improves by 1.1\% $\sim$ 2.1\% AP, indicating that the composed backbone is more efficient than the pre-trained wider and deeper networks. Moreover, CBNet can be flexibly plugged into mainstream detectors (\eg, RetinaNet~\cite{LinGGHD17}, ATSS~\cite{ZhangCYLL20}, Faster R-CNN~\cite{RenHGS15}, Mask R-CNN~\cite{HeGDG17}, Cascade R-CNN and Cascade Mask R-CNN \cite{CaiV18}), and consistently improve the performances of these
detectors by 3\%$\sim$3.8\% AP, demonstrating its strong adaptability to various head designs of detectors. Besides, CBNet is compatible with feature enhancing networks \cite{DBLP:journals/pami/00010CJDZ0MTW0X21, DaiQXLZHW17} and model ensemble method \cite{DBLP:journals/corr/abs-2105-03139}.
Remarkably, it presents a general and resource-friendly framework to drive the accuracy ceiling of high-performance detectors. Without bells and whistles, our \cbname Swin-L achieves unparalleled single-model single-scale result of \ttl{59.4\%} box AP and \ttl{51.6\%} mask AP
on COCO \texttt{test-dev}, surpassing the state-of-the-art result (\textit{i.e.,} 57.7\% box AP and 50.2\% mask AP obtained by Swin-L), while reducing the training schedule by 6$\times$. With multi-scale testing, we push the current best single-model result to a new record of \ttl{60.1\%} box AP and \ttl{52.3\%} mask AP.

The main contributions of this paper are listed as follows:
\begin{itemize}
\item We propose a general, efficient and effective framework, CBNet (Composite Backbone Network), to construct high-performance backbone networks for object detection without additional pre-training. 
\item We propose a Dense Higher-Level Composition (DHLC) strategy, auxiliary supervision, and a pruning strategy to efficiently use existing pre-trained weights for object detection under the pre-training fine-tuning paradigm. 
\item 
Our \cbname Swin-L achieves a new record of single-model single-scale result on COCO at a shorter (by 6$\times$) training schedule than Swin-L. With multi-scale testing, our method achieves the best-known result without extra training data.
\end{itemize}

\CheckRmv{
\section{Related work}
\label{sec:related_work}
\begin{figure*}[t]
\centering
\includegraphics[width=\textwidth]{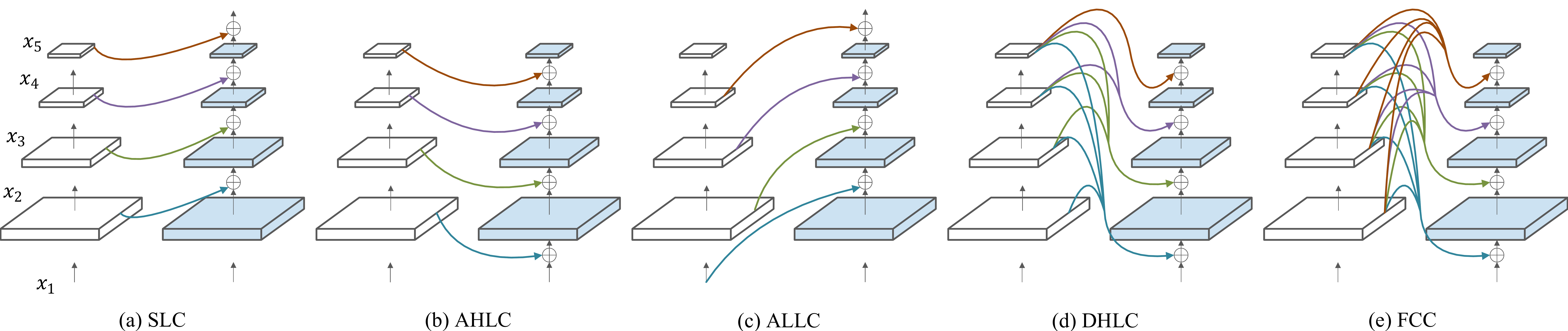}
\caption{Five kinds of composite strategies for Composite Backbone architecture when $K=2$. (a) Same Level Composition (SLC). (b) Adjacent Higher-Level Composition (AHLC). (c) Adjacent Lower-Level Composition (ALLC). (d) Dense Higher-Level Composition (DHLC). (e) Full-connected Composition (FCC). The composite connections are colored lines representing some operations such as element-wise operation, scaling, 1$\times$1 Conv layer, and BN layer. 
}
\label{fig:comp_style}
\end{figure*}

\begin{figure}[t]
\centering
\includegraphics[width=0.9\columnwidth]{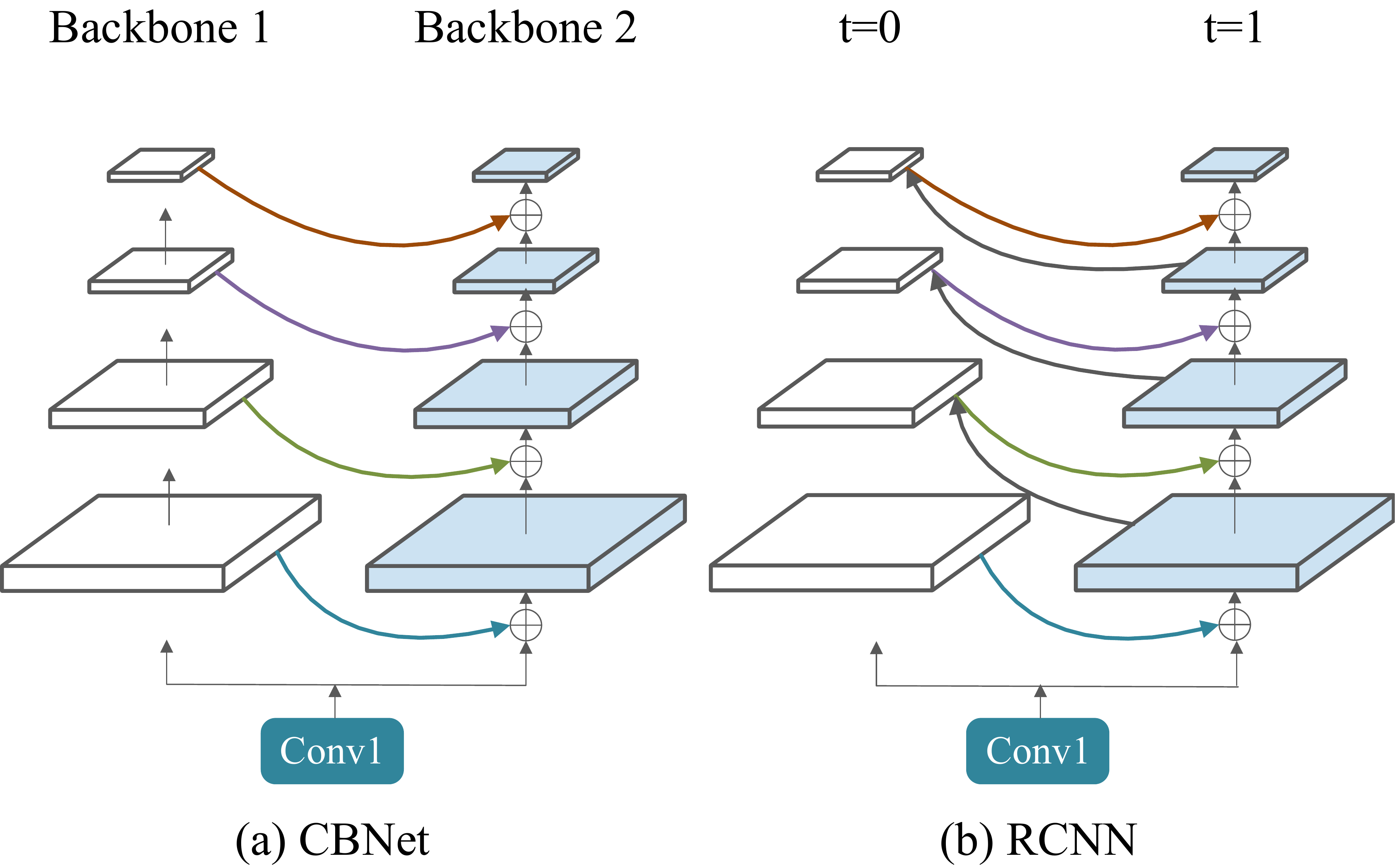}\\
\caption{Comparison between our proposed CBNet architecture ($K=2$) and  the unrolled architecture of RCNN \cite{LiangH15} ($T=2$).}
\label{fig:RCNN}
\vspace{-3mm}
\end{figure}

\ttlv{\subsection{Object Detection}}
Object detection aims to locate each object instance from a predefined set of classes in an input image. With the rapid development of convolutional neural networks (CNNs), there is a popular paradigm for deep learning-based object detectors: the backbone network (typically designed for classification and pre-trained on ImageNet) extracts basic features from the input image, and then the neck (\textit{e.g.}, feature pyramid network~\cite{LinDGHHB17}) enhances the multi-scale features from the backbone, after which the detection head predicts the object bounding boxes with position and classification information. Based on detection heads, the cutting-edge methods for generic object detection can be briefly categorized into two major branches. The first branch contains one-stage detectors such as YOLO~\cite{Redmon_2016_CVPR}, SSD~\cite{LiuAESRFB16}, RetinaNet~\cite{LinGGHD17}, NAS-FPN~\cite{GhiasiLL19},  EfficientDet~\cite{TanPL20}, and \cite{DBLP:journals/corr/abs-2012-01724}. The other branch contains two-stage methods such as Faster R-CNN~\cite{RenHGS15}, FPN~\cite{LinDGHHB17}, Mask R-CNN~\cite{HeGDG17}, Cascade R-CNN~\cite{CaiV18}, and Libra R-CNN~\cite{pang2019libra}. Recently, academic attention has been geared toward anchor-free detectors due partly to the emergence of FPN~\cite{LinDGHHB17} and focal Loss~\cite{LinGGHD17}, where more elegant end-to-end detectors have been proposed. On the one hand, FSAF~\cite{zhu2019feature}, FCOS~\cite{TianSCH19}, ATSS~\cite{ZhangCYLL20} and GFL~\cite{0041WW00LT020} improve RetinaNet with center-based anchor-free methods. On the other hand, CornerNet~\cite{LawD20}, CenterNet~\cite{DuanBXQH019}, and FoveaBox \cite{DBLP:journals/tip/KongSLJLS20} detect object bounding boxes with a keypoint-based method.  In addition to the above CNN-based detectors, Transformer~\cite{VaswaniSPUJGKP17} has also been utilized for detection.  DETR~\cite{CarionMSUKZ20} proposes a fully end-to-end detector by combining CNN and Transformer encoder-decoders.

More recently, Neural Architecture Search (NAS) is applied to automatically search the architecture for a specific detector. NAS-FPN~\cite{GhiasiLL19}, NAS-FCOS~\cite{WangGCWTSZ20} and SpineNet~\cite{DuLJGTCLS20} use reinforcement learning to control the architecture sampling and obtain promising results. SM-NAS~\cite{YaoXZLL20} uses the evolutionary algorithm and partial order pruning method to search the optimal combination of different parts of the detector. Auto-FPN~\cite{XuYLLZ19} uses the gradient-based method to search for the best detector. OPANAS \cite{Liang_2021_CVPR} uses the one-shot method to search for an efficient neck for object detection.

\ttlv{\subsection{Backbones for Object Detection}}
Starting from AlexNet \cite{krizhevsky2012imagenet}, deeper and wider backbones have been exploited by mainstream detectors, such as VGG \cite{simonyan2014very}, ResNet \cite{HeZRS16}, DenseNet  \cite{huang2017densely}, ResNeXt \cite{xie2017aggregated}, and Res2Net \cite{GaoCZZYT21}.
Since the backbone network is usually designed for classification, whether it is pre-trained on ImageNet and fine-tuned on a given detection dataset or trained from scratch on the detection dataset, it requires many computational resources and is difficult to optimize.
Recently, two non-trivially designed backbones, \textit{i.e.}, DetNet \cite{li2018detnet} and FishNet \cite{sun2018fishnet}, are specifically designed for the detection task. However, they still require pre-training for the classification task before fine-tuning for the detection task. Res2Net \cite{GaoCZZYT21} achieves impressive results in object detection by representing multi-scale features at the granular level.
HRNet \cite{DBLP:journals/pami/00010CJDZ0MTW0X21} maintains high-resolution representations and achieves promising results in human pose estimation, semantic segmentation, and object detection.
In addition to manually designing the backbone architecture, DetNAS \cite{ChenYZMXS19} and Joint-DetNAS \cite{YaoPXZLZ21} use NAS to search for a better backbone for object detection, thereby reducing the cost of manual design.  Swin Transformer \cite{abs-2103-14030} and PVT \cite{WangX0FSLL0021} utilize Transformer modular to build the backbone and achieve impressive results, despite the need for expensive pre-training. 

It is well known that designing and pre-training a new and robust backbone requires significant computational costs. Alternatively, we propose a more economical and efficient solution to build a more powerful object detection backbone, by grouping multiple identical existing backbones (\textit{e.g.}, ResNet  \cite{HeZRS16}, ResNeXt \cite{xie2017aggregated}, Res2Net  \cite{GaoCZZYT21}, HRNet \cite{DBLP:journals/pami/00010CJDZ0MTW0X21}, and Swin Transformer \cite{abs-2103-14030}).

\ttlv{\subsection{Recurrent Convolution Neural Network}}
Different from the feed-forward architecture of CNN, Recurrent CNN (RCNN) \cite{LiangH15} incorporates recurrent connections into each convolution layer to enhance the contextual information integration ability of the model.
As shown in \figref{fig:RCNN}, our proposed  Composite Backbone Network shares some similarities with the unfolded RCNN~\cite{LiangH15}, but they are very different. First, the connections between the parallel stages in CBNet are unidirectional, while they are bidirectional in RCNN. Second, in RCNN, the parallel stages at different time steps share parameter weights, while in the proposed CBNet, the parallel stages of backbones are independent of each other. Moreover, we need to pre-train RCNN on ImageNet if we use it as the backbone of the detector. By contrast, CBNet does not require additional pre-training because it directly uses existing pre-trained weights.

\ttlv{\subsection{Model Ensemble}}
It is well known that a combination of many different predictors can lead to more accurate predictions, \eg, 
ensemble methods are considered \ttl{as} the state-of-the-art solution for many machine learning challenges. The model ensemble improves the prediction performance of a single model by training multiple different models and combining their prediction results through post processing\cite{nips/KroghV94, widm/SagiR18}.

There are two key characteristics for model ensemble: \textit{model diversity} and \textit{voting}.
Model diversity means that the models with different architectures or training techniques are trained separately, and its importance for the model ensemble is well established  \cite{book/ensem_ml, ethos/Brown04a, inffus/BrownWHY05, ChenFL21cvpr}. Most ensemble methods need voting strategies to compare the outputs of different models and refine the final predictions  \cite{DBLP:journals/corr/abs-2105-03139}. In terms of the above two characteristics, our CBNet is very different from the model ensemble. In fact, CBNet benefits from the \textit{identical} backbones grouping, the recurrent style feature enhancing by  \textit{jointly} training. Furthermore, the output of the lead backbone is used directly for the final prediction without the need to be assembled with other backbones. More practical analysis can be found in \secref{sec:ensem}.

In practice, leading approaches to the challenge object detection benchmarks like  MS COCO  \cite{LinMBHPRDZ14} or OpenImage \cite{DBLP:journals/ijcv/KuznetsovaRAUKP20} are based on the usage of model ensemble \cite{DBLP:journals/pami/RenHGZS17, DBLP:conf/cvpr/HuangRSZKFFWSG017, peng2018megdet, DBLP:conf/cvpr/LiuQQSJ18, DBLP:journals/corr/abs-2010-02475, DBLP:journals/corr/abs-2003-07557}. For example, \cite{DBLP:journals/corr/abs-2003-07557} separately trains 28 models of different architectures, heads, data splits, class sampling strategies, augmentation strategies and supervisions and aggregate these detector's outputs by ensembling method. \cite{DBLP:journals/corr/abs-2105-03139} proposes the Probabilistic Ranking Aware Ensemble (PRAE) that refines the confidence of bounding boxes from different detectors. Our CBNet is compatible with such model ensemble methods, as are other conventional backbones. More details can be found in \secref{sec:identical}.

\ttlv{\subsection{Our Approach.~}}
Our network groups multiple identical backbones in parallel. It integrates the high- and low-level features of multiple identical backbones and gradually expands the receptive field to more efficiently perform object detection. This paper represents a very substantial extension of our previous conference paper \cite{LiuWWLZTL20} with results under recently developed start-of-the-art object detection frameworks. The main technical novelties compared with \cite{LiuWWLZTL20} lie in three aspects. (1) We extend the network (named as CBNetV1) proposed in \cite{LiuWWLZTL20}, with three modifications: a specialized training method, a better composite strategy and a pruning strategy, which respectively optimizes the training process, more efficiently enhances feature representation and reduces the model complexity of CBNetV2. (2) We show the strong generalization capabilities of CBNetV2 for various backbones and head designs of detector architecture. (3) We show the superiority of CBNetV2 over CBNetV1 and present the state-of-art result of CBNetV2 in object detection.

}
\CheckRmv{
\section{Proposed method}
\label{sec:method}
This section elaborates the proposed CBNet in details. In \secref{sec:arc} and \secref{sec:opcab}, we describe its basic architecture and variants, respectively. In \secref{sec:supervision}, we propose a training strategy for CBNet-based detectors. In \secref{sec:pruning}, we briefly introduce the pruning strategy. In \secref{sec:det}, we summarize the detection framework of CBNet.

\subsection{Architecture of CBNet}
\label{sec:arc}

The proposed CBNet consists of $K$ identical backbones ($K\ge2$). 
In particular, we call the case $K=n$ as \cbname Backbone-K$n$, where '-K$n$' is omitted when $K=2$.

As in \figref{fig:shouye}, the CBNet architecture includes two types of backbones: lead backbone $B_{K}$ and assisting backbones $B_{1}, B_{2},... , B_{K-1}$. Each backbone comprises $L$ stages (usually $L=5$), and each stage consists of several convolutional layers with feature maps of the same size. The $l$-th stage of the backbone implements the non-linear transformation $F^l(\cdot) (l=1,2,...,L)$.

Most conventional convolutional networks follow the design of encoding the input images into intermediate features with monotonically decreased resolution. 
In particular, the $l$-th stage takes the output (denoted as $x^{l-1}$) of the previous ($l-1$)-th stage as input, which can be expressed as follows:
\begin{equation}
\label{equa1}
x^l = F^l(x^{l-1}),  l\ge2.
\end{equation}
Differently, we adopt assisting backbones $B_{1}, B_{2},... , B_{K-1}$ to improve the representative ability of lead backbone $B_K$. We iterate the features of a backbone to its successor in a stage-by-stage fashion. 
Thus, \eqref{equa1} can be rewritten as:
\begin{equation}
x_k^l = F_k^l(x_k^{l-1} + g^{l-1}(\boldsymbol{x_{k-1}})),  ~~  l\ge2,k=2,3, \ldots,K, 
\label{eq:cbv2}
\end{equation}
where $g^{l-1}(\cdot)$ represents the composite connection, which takes features (denoted as $\boldsymbol{x_{k-1}} = \{x_{k-1}^i|i=1,2,\ldots, L\}$) from assisting backbone $B_{k-1}$ as input and takes the features of the same size as $x_k^{l-1}$ as output. 
Therefore, the output features of $B_{k-1}$ are transformed and contribute to the input of each stage in $B_k$. Note that $x_1^1, x_2^1, \ldots, x_K^1$ are weight sharing.

For the object detection task, only the output features of  the lead backbone $\{x_K^{i}, i = 2,3,\ldots,L\}$ are fed into the neck and then the RPN/detection head, while the outputs of the assisting backbone are forwarded to its succeeding siblings. It is worth noting that $B_{1}, B_{2},... , B_{K-1}$ can be used for various backbone architectures (\textit{e.g.,} ResNet  \cite{HeZRS16}, ResNeXt \cite{xie2017aggregated}, Res2Net  \cite{GaoCZZYT21}, and Swin Transformer \cite{abs-2103-14030}) and initialized directly from the pre-trained weights of a single backbone.

\CheckRmv{
\begin{figure}[t!]
\centering
\includegraphics[width=\columnwidth]{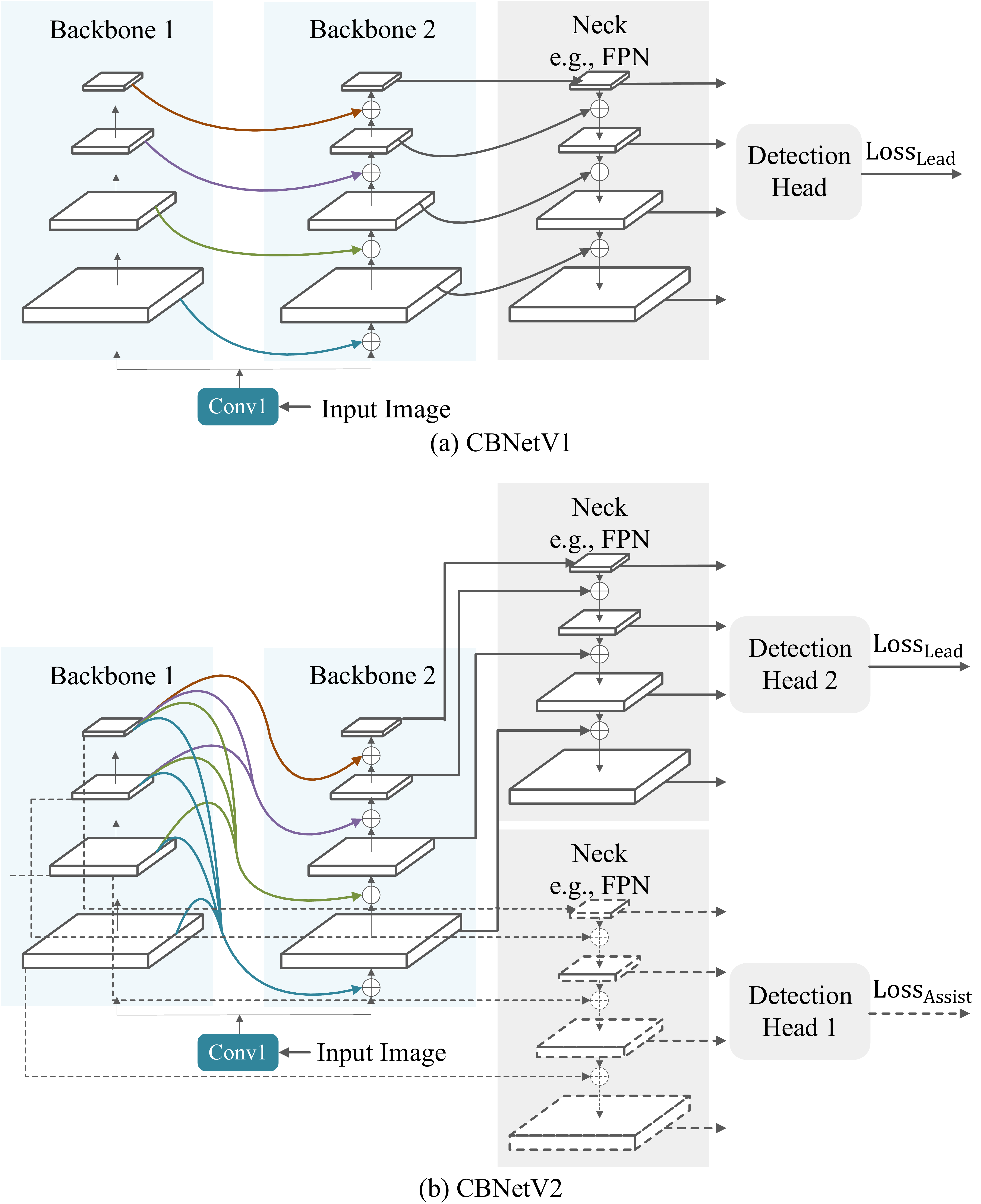}
\caption{(a) CBNetV1 \cite{LiuWWLZTL20} ($K=2$). (b) CBNetV2 ($K=2$) with auxiliary supervision. The two FPNs and detection heads share the same weights. It is worth noting that the main differences between CBNetV2 and CBNetV1 are in composite strategies and training strategies.
}
\label{fig:supervise_db}
\end{figure}
}
\subsection{Possible Composite Strategies}
\label{sec:opcab}
For composite connection $ g^l(x)$ which takes $\boldsymbol{x}=\{x^i| i=1,2,\ldots,L\}$ from an assisting backbone as input and outputs a feature of the same size of $x^l$ (omitting $k$ for simplicity), we propose the following five different composite strategies.
\CheckRmv{
\begin{figure*}[!t]
\centering
\includegraphics[width=\textwidth]{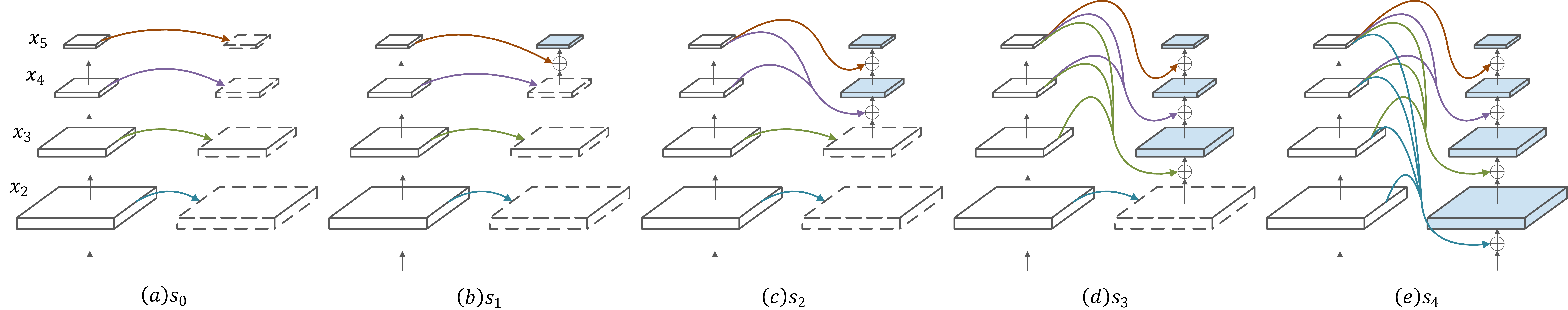}
\caption{
Illustration of different pruning strategies for composite backbone when $K=2$.  
 \ttl{$s_i$ indicates there are $i$ stages $\{x_j|j\geq6-i  {\rm~and~} j\leq5, i=0,1,2,3,4$\}
in the $2, 3, ..., K$-th backbone and the pruned stages are filled by the features of the same stages in the first  backbone.}
}
\label{fig:simpcb}
\end{figure*}
}

\subsubsection{Same Level Composition (SLC)} 
An intuitive and simple way of compositing is to fuse the output features from the same stage of backbones. As shown in \figref{fig:comp_style}.a, the operation of SLC can be formulated as:
\begin{equation}
g^l(x)= \mathbf{w}(x^l),  l\ge2,
\end{equation}
where $\mathbf{w}$ represents a $1\times 1$ convolution layer and a  batch normalization layer.

\subsubsection{Adjacent Higher-Level Composition (AHLC)} 
\label{sec:ahlc}
Motivated by Feature Pyramid Networks \cite{LinDGHHB17}, the top-down pathway introduces the spatially coarser, but semantically stronger, higher-level features to enhance the lower-level features of the bottom-up pathway,
we introduce AHLC to feed the output of the adjacent higher-level stage of the previous backbone to the subsequent one (from left to right in \figref{fig:comp_style}.b):
\begin{equation}
g^l(x) = \mathbf{U}(\mathbf{w}(x^{l+1})),  l\ge1,
\end{equation}
where $ \mathbf{U}(\cdot)$ indicates the up-sampling operation. 
\subsubsection{Adjacent Lower-Level Composition (ALLC)} 
Contrary to AHLC, we introduce a bottom-up pathway to feed the output of the adjacent lower-level stage of the previous backbone to the succeeding one. We show ALLC in \figref{fig:comp_style}.c, which is formulated as:
\begin{equation}
g^l(x) =\mathbf{D} (\mathbf{w}(x^{l-1})),  l\ge2,
\end{equation}
where $ \mathbf{D}(\cdot)$ denotes the down-sample operation.

\subsubsection{Dense Higher-Level Composition (DHLC)} 
In DenseNet~\cite{huang2017densely}, each layer is connected to all subsequent layers to build comprehensive features. Inspired by this, we utilize dense composite connections in our CBNet architecture. The operation of DHLC is expressed as follows:
\begin{equation}
g^l(x) =  \sum_{i=l+1}^L \mathbf{U}(\mathbf{w}_i(x^i)),  ~~l\ge1.
\end{equation}
As shown in \figref{fig:comp_style}.d, when $K=2$,
we compose the features from all the higher-level stages in the previous backbone and add them to the lower-level stages in the latter one.

\subsubsection{Full-connected Composition (FCC)} 
As shown in \figref{fig:comp_style}.e, we compose features from all the stages in the previous backbones and feed them to each stage in the following one. Compared to DHLC, we add connections in low-high-level case. The operation of FCC can be expressed as:
\begin{equation}
g^l(x) =  \sum_{i=2}^L \mathbf{I}(\mathbf{w}_i(x^i)),  ~~l\ge1,
\end{equation}
where $\mathbf{I}(\cdot)$ denotes scale-resizing, $\mathbf{I}(\cdot)= \mathbf{D}(\cdot)$ when $i>l$, and $\mathbf{I}(\cdot)= \mathbf{U}(\cdot)$ when $i<l$.

\subsection{Auxiliary Supervision}
\label{sec:supervision}
Although increasing the depth usually leads to performance improvement~\cite{HeZRS16}, it may introduce additional optimization difficulties, as in the case of image classification \cite{ShenLH16}. The studies in~\cite{SzegedyLJSRAEVR15, SzegedyVISW16} introduce the auxiliary classifiers of intermediate layers to improve the convergence of very deep networks.
In original CBNet, although the composite backbones are parallel, the latter backbone (\textit{e.g.,} lead backbone in \figref{fig:supervise_db}.a) deepen the network through adjacent connections between the previous backbone (\textit{e.g.,} assisting backbone in \figref{fig:supervise_db}.a). 
To better train the CBNet-based detector, We propose to generate initial results of assisting backbones by supervision with the auxiliary neck and detection head to provide additional regularization.

An example of our supervised CBNet when $K$=2 is illustrated in \figref{fig:supervise_db}.b. Apart from the original loss that uses the lead backbone feature to train the detection head 1, another detection head 2 takes assisting backbone features as input, producing \textit{auxiliary supervision}. Note that detection head 1 and detection head 2 are weight sharing, as are the two necks.  The auxiliary supervision helps to optimize the learning process, while the original loss for the lead backbone takes the greatest responsibility. We add weights to balance the auxiliary supervision, where the total loss is defined as:
\begin{equation}
\label{eq:supervision}
\mathcal{L} = \mathcal{L}_\mathrm{Lead} + \sum_{i=1}^{K-1}(\lambda_i\cdot \mathcal{L}_\mathrm{Assist}^i).
\end{equation}
where $\mathcal{L}_\mathrm{Lead}$ is the loss of lead backbone,  $\mathcal{L}_\mathrm{Assist}$ is the loss of assisting backbones, and $\lambda_i$ is the loss weight for the $i$-th assisting backbone.

During the inference phase, we abandon the auxiliary supervision branch and only utilize the output features of the lead backbone in CBNet (\figref{fig:supervise_db}.b).   
Consequently, auxiliary supervision does not affect the inference speed.

\subsection{Pruning Strategy for CBNet }
\label{sec:pruning}
To reduce the model complexity of CBNet, we explore the possibility of pruning the different number of stages in  $2, 3, ..., K$-th backbones instead of composing the backbones in a holistic manner. For simplicity, we show five pruning methods when $K=2$ in \figref{fig:simpcb}.  \ttl{$s_i$ indicates there are $i$ stages $\{x_j|j\geq6-i  {\rm~and~} j\leq5, i=0,1,2,3,4$\}
in the $2, 3, ..., K$-th backbone} and the pruned stages are filled by the features of the same stages in the first  backbone. 

\subsection{Architecture of Detection Network with CBNet }
\label{sec:det}
CBNet can be applied to various off-the-shelf detectors without additional modifications to network architectures.
In practice, we attach the lead backbone with functional networks, \eg, FPN \cite{LinDGHHB17} and detection head. The inference phase of CBNet is shown in \figref{fig:shouye}. Note that we present two versions of CBNet. The first one, named CBNetV1 \cite{LiuWWLZTL20}, uses only AHLC composition strategy, providing a simple implementation of the composite backbone that is easy to follow. The other one, CBNetV2, combines DHLC composition strategy, the auxiliary supervision, and a special pruning strategy, to fully explore the potential of CBNet for object detection. We empirically demonstrate the superior of CBNetV2 over CBNetV1 in the following \secref{sec:exp}. In this paper, CBNet denotes CBNetV2 in the following experiments if not specified.

}
\CheckRmv{\section{Experiments}
\label{sec:exp}
\CheckRmv{
\begin{table*}[t!]
\tabFormat
\setlength{\tabcolsep}{2.3mm}
\caption{Comparison with the state-of-the-art results on COCO \texttt{test-dev}.
Our \cbname Res2Net101-DCN achieves higher bbox AP over previous anchor-free and anchor-based detectors while using comparable or fewer training epochs.}
\begin{threeparttable}
\begin{tabular}{c|c|c|ccccc|c|c}
\toprule
& \textbf{Method}   & $\mathbf{AP^{box}}$ & $\mathbf{AP^{box}_{50}}$ & $\mathbf{AP^{box}_{75}}$ & $\mathbf{AP^{box}_{S}}$ & $\mathbf{AP^{box}_{M}}$ & $\mathbf{AP^{box}_{L}}$ & \textbf{Params} & \textbf{Epochs}   \\ \hline
\parbox[t]{5pt}{\multirow{9}{*}{\rotatebox[origin=c]{90}{\textit{\textbf{anchor-free-based}}}}} 
& FSAF \cite{zhu2019feature}   & 42.9 &63.8 &46.3 &26.6 &46.2 &52.7  & 94 M  & 24   \\ 
& FCOS \cite{TianSCH19}  & 44.7 & 64.1   & 48.4   & 27.6  & 47.5  & 55.6  & 90 M  & 24   \\ 
& DETR-DC5 \cite{CarionMSUKZ20} & 44.9 & 64.7   & 47.7   & 23.7  & 49.5  & 62.3  & 60 M   & 500  \\ 
& NAS-FCOS \cite{WangGCWTSZ20}  & 46.1 & -    & -    & -   & -   & -   & 89 M  & 24   \\ 
& ATSS \cite{ZhangCYLL20}   & 47.7 &66.5 &51.9 &29.7 &50.8 &59.4  & 95 M  & 24   \\ 
& GFL \cite{0041WW00LT020}    & 48.2 & 67.4   & 52.6   & 29.2  & 51.7  & 60.2  & 53 M  & 24   \\ 
& Deformable DETR \cite{zhu2020deformable} &50.1 &69.7 &54.6 &30.6 &52.8 &64.7  & -   & 50 \\ 
& PAA \cite{KimL20} &50.8 &69.7 &55.1 &31.4 &54.7 &65.2 &129 M &24\\
& \textbf{\cbname Res2Net101-DCN (ATSS)}   & \textbf{52.7}  & \textbf{71.1}  & \textbf{57.7}  & \textbf{35.7}    & \textbf{56.6}    & \textbf{62.7}    & \textbf{107 M}  & \textbf{20} \\ \hline
\parbox[t]{5pt}{\multirow{9}{*}{\rotatebox[origin=c]{90}{\textit{\textbf{anchor-based}}}}} 
& Auto-FPN \cite{XuYLLZ19}    & 44.3 & -    & -    & -   & -   & -   & 90 M  & 24   \\ 
& SM-NAS:E5 \cite{YaoXZLL20}  & 45.9 & 64.6   & 49.6   & 27.1  & 49.0  & 58.0  & -   & 24   \\ 
& NAS-FPN \cite{GhiasiLL19}   & 48.3 & -    & -    & -   & -   & -   & -   & 150  \\ 
& SP-NAS \cite{JiangXZLL20}   & 49.1 & 67.1   & 53.5   & 31  & 52.6  & 63.7  & -   & 50   \\ 
& \ttl{Joint-DetNAS} \cite{YaoPXZLZ21}&\ttl{50.7} &\ttl{69.6} &\ttl{55.4} &\ttl{31.3} &\ttl{53.8} &\ttl{64.0}  & \ttl{-}   & \ttl{16} \\ 
& SpineNet-190 \cite{DuLJGTCLS20}  & 52.1 & 71.8   & 56.5   & 35.4  & 55  & 63.6  & 164 M  & 250  \\ 
& OPANAS  \cite{Liang_2021_CVPR} &52.2 &71.3 &57.3 &33.3 &55.6 &65.4 &83 M &24\\
& EfficientDet-D7x \cite{TanPL20}  & 55.1 & 74.3   & 59.9   & -   & -   & -   & 77 M   & 600  \\ 
& YOLOv4-P7  \cite{Wang_2021_CVPR} & 55.5 & 73.4   & 60.8   & 38.4  & 59.4  & 67.7  & 288 M  & 450  \\  		
 & \textbf{\cbname Res2Net101-DCN (Cascade R-CNN)}  & \textbf{55.6}  & \textbf{73.7}  & \textbf{60.8}  & \textbf{37.4}    & \textbf{59.0}    & \textbf{67.6}    & \textbf{146 M}  & \textbf{32} \\ 		
\bottomrule
\end{tabular}
\end{threeparttable}
\label{tab:sota}
\end{table*}
}

\CheckRmv{
\begin{table*}[t!]
\tabFormat
\setlength{\tabcolsep}{2.3mm}
\caption{Comparison with the state-of-the-art object detection and instance segmentation results on COCO.  In collaboration with Swin Transformer, our CBNetV2 achieves the state-of-the-art box AP and mask AP while using fewer training epochs.
}
\begin{threeparttable}

\begin{tabular}{c|c|cc|cc|c|c}
\toprule
\multirow{2}{*}{\textbf{Pre-trained   on}} & \multirow{2}{*}{\textbf{Method}}  & \multicolumn{2}{c|}{\textbf{mini-val}}  & \multicolumn{2}{c|}{\textbf{test-dev}}  & \multirow{2}{*}{\textbf{Params}} & \multirow{2}{*}{\textbf{Epochs}} \\ 
&   & $\mathbf{AP^{box}}$ & $\mathbf{AP^{mask}}$ & $\mathbf{AP^{box}}$ & $\mathbf{AP^{mask}}$ &  &   \\ \hline
\multirow{5}{*}{ImageNet-1K} & GCNet$^\ast$ \cite{9307278}  & 51.8& 44.7 & 52.3& 45.4 & -& 36\\
& Swin-B~\cite{abs-2103-14030} & 51.9& 45.0 & -   & - & 145 M  & 36\\
& ResNeSt-200$^\ast$   \cite{abs-2004-08955}   & 52.5& - & 53.3& 47.1 & -& 36\\
& CopyPaste~\cite{abs-2012-07177}$^\dagger$   & 55.9& 47.2 & 56.0& 47.4 & 185 M  & 96\\

& \textbf{\cbname Swin-S (Cascade Mask   R-CNN)}  & \textbf{56.3}& \textbf{48.6}& \textbf{56.9}& \textbf{49.1}& \textbf{156 M} & \textbf{36}\\ \hline
\multirow{6}{*}{ImageNet-22K}  & Swin-L (HTC++)~\cite{abs-2103-14030} & 57.1& 49.5 & 57.7& 50.2 & 284 M  & 72\\
& Swin-L   (HTC++)~\cite{abs-2103-14030}$^\ast$   & 58.0& 50.4 & 58.7& 51.1 & 284 M  & 72\\
& \cbname Swin-B (HTC)                   &58.4       & 50.7        & 58.7       &51.1        & 235 M                   & 20                      \\
& \cbname Swin-B (HTC)$^\ast$             & 58.9       &51.3        & 59.3       & 51.8        & 235 M                   & 20                    \\
& \cbname Swin-L (HTC)                    & 59.1      & 51.0        & 59.4       & 51.6        & 453 M                   &12                     \\

& \textbf{\cbname Swin-L (HTC)$^\ast$}             & \textbf{59.6}       & \textbf{51.8}        & \textbf{60.1}       & \textbf{52.3}        & \textbf{453 M}                   & \textbf{12}   \\
\bottomrule
\end{tabular}
 \begin{tablenotes}
       \footnotesize
       \item[1] $^\ast$ indicates method with multi-scale testing.
       \item[2] $^\dagger$ indicates additional unlabeled data are used to pretrain backbone.
\end{tablenotes}
\end{threeparttable}
\label{tab:sotaswin}
\end{table*}}

In this section, we evaluate our CBNet through extensive experiments. In \secref{sec:implement}, we detail the experimental setup. In \secref{sec:exp_sota},  we compare CBNet with state-of-the-art detection methods. In \secref{sec:general}, we demonstrate the generality of our method over different backbones and detectors. In \secref{sec:compatible}, we show the compatibility of CBNet with DCN and model ensemble. In \secref{sec:ablation}, we conduct an extensive ablation study to investigate individual components of our framework.

\subsection{Implementation details}
\label{sec:implement}

\subsubsection{Datasets and Evaluation Criteria}
We conduct experiments on the COCO \cite{LinMBHPRDZ14} benchmark. The training is conducted on the 118k training images, and ablation studies on the 5k \texttt{minival} images. We also report the results on the 20k images in \texttt{test-dev} for comparison with the state-of-the-art  (SOTA) methods. For evaluation, we adopt the metrics from the COCO detection evaluation criteria, including the mean Average Precision  (AP) across IoU thresholds ranging from 0.5 to 0.95 at different scales. 

\subsubsection{Training and Inference Details}
Our experiments are based on the open-source detection toolbox  MMDetection \cite{mmdetection}. For ablation studies and simple comparisons, we resize the input size to $800\times 500$ during training and inference if not specified. We choose Faster R-CNN  (ResNet50 \cite{HeZRS16}) with FPN \cite{LinDGHHB17} as the baseline. We use the SGD optimizer with an initial learning rate of 0.02, the momentum of 0.9, and $10^{-4}$ as weight decay. We train detectors for 12 epochs with a learning rate decreased by $10\times$ at epoch 8 and 11. We use only random flip for data augmentation and set the batch size to 16. 
Note that experiments related to \ttl{special backbones, e.g., Swin Transformer \cite{abs-2103-14030,SandlerHZZC18}, HRNet \cite{DBLP:journals/pami/00010CJDZ0MTW0X21}, PVT \cite{WangX0FSLL0021}, PVTv2~\cite{wang2022pvt}, and DetectoRS \cite{QiaoCY21},}
are not highlighted specifically following the hyper-parameters of \ttl{the original papers}.
The inference speed FPS (frames per second) for the detector is measured on a machine with 1 V100 GPU.

To compare with state-of-the-art detectors, we utilize multi-scale training \cite{ChenPWXLSF0SOLL19}  (the short side resized to $400\sim 1400$ and the long side is at most 1600) and a longer training schedule (details can be found in \secref{sec:exp_sota}). During the inference phase, we use Soft-NMS \cite{bodla2017soft} with a threshold of 0.001, and the input size is set to $1600\times 1400$. All other hyper-parameters in this paper follow MMDetection if not specified.

\subsection{Comparison with State-of-the-Art}
\label{sec:exp_sota}
We compare our methods with cutting-edge detectors. We divide the results into object detection (\tabref{tab:sota}) and instance segmentation (\tabref{tab:sotaswin}) according to whether or not the instance segmentation annotations are used during training. Following \cite{abs-2103-14030},  we improve the detector heads of Cascade R-CNN, Cascade Mask R-CNN, and HTC in the above two tables by adding four convolution layers \cite{ijcv/WuH20} in each bounding box head and using GIoU loss \cite{RezatofighiTGS019} instead of Smooth L1 \cite{girshick2015fast}.
\subsubsection{Object Detection}
For detectors trained with only bounding box annotations, we summarize them into two categories: anchor-based, and anchor-free-based in \tabref{tab:sota}. We select ATSS \cite{ZhangCYLL20} as the anchor-free representative, and Cascade R-CNN as the anchor-based representative. 

\vspace{1mm}\noindent\textbf{Anchor-free.~}
\cbname Res2Net101-DCN equipped with ATSS is trained for 20 epochs, where the learning rate is decayed by $10\times$ in the 16th and 19th epochs. Notably, our \cbname Res2Net101-DCN achieves 52.8\% AP, outperforming previous anchor-free methods \cite{zhu2019feature, TianSCH19, CarionMSUKZ20, WangGCWTSZ20, 0041WW00LT020, ZhangCYLL20, zhu2020deformable} under single-scale testing protocol. 

\vspace{1mm}\noindent\textbf{Anchor-based.~}
Our \cbname Res2Net101-DCN achieves 55.6\% AP, surpassing other anchor-based detectors  \cite{XuYLLZ19, YaoXZLL20, GhiasiLL19, JiangXZLL20, DuLJGTCLS20, Dai_2021_CVPR, TanPL20, Liang_2021_CVPR}. It is worth noting that our CBNet trains only for 32 epochs (the first 20 epochs are regular training and the remaining 12 epochs are trained with Stochastic Weights Averaging~\cite{zhang2020swa}), being 16$\times$ and 12$\times$ shorter than EfficientDet and YOLOv4, respectively.

\subsubsection{Instance Segmentation}
We further compare our method with state-of-the-art results \cite{9307278, abs-2004-08955, abs-2012-07177, abs-2103-14030} using both bounding box and instance segmentation annotations in \tabref{tab:sotaswin}. Following \cite{abs-2103-14030}, we provide results with the backbone pre-trained on regular ImageNet-1K and ImageNet-22K to show the high capacity of CBNet.

\vspace{1mm}\noindent\textbf{Results with regular ImageNet-1K pre-train.~} 
Following~\cite{abs-2103-14030}, 3x schedule  (36 epochs with the learning rate decayed by $10\times$ at epochs 27 and 33) is used for \cbname Swin-S. Using Cascade Mask R-CNN, our \cbname Swin-S achieves 56.3\%  box AP and 48.6\% mask AP on COCO \texttt{minival} in terms of the bounding box and instance segmentation, showing significant gains of +4.4\% box AP and +3.6\% mask AP to Swin-B with similar model size and the same training protocol. In addition, \cbname Swin-S achieves 56.9\% box AP and 49.1\% mask AP on COCO \texttt{dev}, outperforming other ImageNet-1K pre-trained backbone-based detectors. 

\vspace{1mm}\noindent\textbf{Results with ImageNet-22K pre-train.~}
 Our \cbname Swin-B achieves single-scale result of 58.4\% box AP and 50.7\% mask AP on COCO \texttt{minival}, which is  1.3\% box AP and 1.2\% mask AP higher than that of Swin-L (HTC++)~\cite{abs-2103-14030} while the number of parameters is decreased by 17\% and the training schedule is reduced by 3.6$\times$.
Especially, with only 12 epochs training (which is 6$\times$ shorter than Swin-L), our \cbname Swin-L achieves 59.4\% box AP and 51.6\% mask AP on COCO \texttt{test-dev}, outperforming prior arts. We can push the current best result to a new record of 60.1\% box AP and 52.3\% mask AP through multi-scale testing.  The results demonstrate that CBNet proposes an efficient, effective, and resource-friendly framework to build high-performance detectors.

\subsection{Generalization Capability of CBNet}
\label{sec:general}
CBNet expands the receptive field by combining the backbones in parallel rather than simply increasing the depth of the network. 
To demonstrate the generality of our design strategy, we perform experiments on various backbones and different head designs of the detector architecture.

\CheckRmv{
\begin{table}[t!]
\tabFormat
\caption{Comparison between CBNet and conventional backbones in terms of different  architectures.   Our CBNet boosts the performance of  CNN-based backbones by over 3.4\% AP.
} 
\vspace{-2mm}
\begin{tabular}{c|ccc} 
\toprule
\multirow{2}{*}{\textbf{Backbone}} & \multicolumn{2}{c}{$\mathbf{AP^{box}}$} & \multirow{2}{*}{$\Delta\mathbf{AP}$} \\
& Origin   & \cbname   & \\ \hline
ResNet50 & 34.6 & \textbf{38.0}  & +3.4 \\
ResNeXt50-32x4d & 36.3 & \textbf{39.8}  & +3.5 \\
Res2Net50  & 37.7 & \textbf{41.2}  & +3.5 \\
\bottomrule
\end{tabular}
\label{tab:50db} 
\end{table}
}
\CheckRmv{
\begin{table}[t!]
\tabFormat
\setlength{\tabcolsep}{2.3mm}
\caption{Comparison between \cbname Backbone 
and deeper and wider single backbones in terms of different architectures. The backbones in each group are sorted by FLOPs for efficiency comparison. Backbones armed with our proposed method are more efficient than their own wider and deeper version.}
\vspace{-2mm}
\begin{tabular}{ccccc} 
\toprule
\textbf{Backbone}   & $\mathbf{AP^{box}}$ &  \textbf{Params} &\textbf{FLOPs} &\textbf{FPS}\\ 
\hline
ResNet101  & 36.3  & 60.5 M &121 G &25.8 \\
\textbf{\cbname ResNet50}  &\textbf{38.0} &\textbf{69.4 M} &\textbf{121 G}&\textbf{23.3}\\
ResNet152  & 37.8  & 76.2 M &151 G &21.3\\

\hline
ResNeXt101-32x4d  &37.7  & 60.2 M &122 G &20.8\\
\textbf{\cbname ResNeXt50-32x4d}  &\textbf{39.8}& \textbf{68.4 M} &\textbf{123 G} &\textbf{19.3}\\
ResNeXt101-64x4d  & 38.7  & 99.3 M &183 G &15.8\\

\hline
Res2Net101 &40.1 & 61.2 M & 125 G &20.7\\
\textbf{\cbname Res2Net50}  &\textbf{41.2} &\textbf{69.7 M} &\textbf{125 G} &\textbf{20.2}\\
Res2Net200 &41.7 & 92.2 M & 185 G  &14.8\\

\bottomrule
\end{tabular}
\label{tab:dbcomparedeep} 
\end{table}
}
\subsubsection{Generality for Main-stream Backbone Architectures}
\label{sec:exp_backbone}
\vspace{1mm}\noindent\textbf{Effectiveness.~} To demonstrate the effectiveness of CBNet, we conduct experiments on Faster R-CNN with different backbone architectures. As shown in \tabref{tab:50db}, for CNN-based backbones (\textit{e.g.,} ResNet, ResNeXt-32x4d, and Res2Net), our method can boost baseline by over 3.4\% AP.

\vspace{1mm}\noindent\textbf{Efficiency.~} Note that the number of parameters in CBNet has increased compared to the baseline. To better demonstrate the efficiency of the composite architecture, we compare CBNet with deeper and wider backbone networks. As shown in \tabref{tab:dbcomparedeep},  with comparable number of number and inference speed, CBNet improves ResNet101, ResNeXt101-32x4d, Res2Net101 by 1.7\%, 2.1\%, and 1.1\% AP, respectively.  Additionally, \cbname ResNeXt50-32x4d is 1.1\% AP higher than that of ResNeXt101-64x4d, while the number of parameters is only 70\%.  
The results demonstrate that our composite backbone architecture is more efficient and effective than simply increasing the depth and width of the network.

\subsubsection{Generality for Swin Transformer}
\label{sec:swin}
Transformer is notable for the use of attention to model long-range dependencies in data, and Swin Transformer \cite{abs-2103-14030} is one of the most representative recent arts. 
We conduct experiments on Swin Transformer to show the model generality of CBNet. For a fair comparison, we follow the same training strategy as \cite{abs-2103-14030} with multi-scale training  (the short side resized to $480\sim 800$ and the long side at most 1333), AdamW optimizer  (initial learning rate of 0.0001, weight decay of 0.05, and batch size of 16), and 3x schedule (36 epochs). 
As shown in \tabref{tab:transform}, the accuracy of the model slowly increases as the Swin Transformer is deepened and widened, and saturates at Swin-S.  Swin-B is only 0.1\% AP higher than that of Swin-S, but the amount of parameters increases by 38M. When using \cbname Swin-T, we achieve 53.6\% box AP and 46.2\% mask AP by improving Swin-T 3.1\% box AP and 2.5\% mask AP. Surprisingly, our \cbname Swin-T is 1.7\% box AP and 1.2\% mask AP higher than that of the deeper and wider Swin-B while the model complexity is lower (\textit{e.g.,} FLOPs 836G vs. 975G, Params 113.8M vs. 145.0M, FPS 6.5 vs. 5.9). These results prove that CBNet can also improve non-pure convolutional architectures. They also demonstrate that CBNet pushes the upper limit of accuracy for high-performance detectors more effectively than simply increasing the depth and width of the network.
\CheckRmv{
\begin{table}[t!]
\tabFormat
\caption{Comparison between \cbname Swin-T and the Swin equipped with Cascade Mask R-CNN. CBNet is more effective and efficient than the wider and deeper version of Swin-Transformer.}
\vspace{-2mm}
\begin{tabular}{cccccc} 
\toprule
\textbf{Backbone}   & $\mathbf{AP^{box}}$  &$\mathbf{AP^{mask}}$  &\textbf{Params} &\textbf{FLOPs}& \textbf{FPS}\\ 
\hline
Swin-T  &50.5 &43.7& 85.6 M &742 G & 7.8\\
Swin-S  &$51.8^{+1.3}$ &$44.7^{+1.0}$& 107.0 M & 832 G &7.0\\
Swin-B  &$51.9^{+1.4}$ &$45.0^{+1.3}$& 145.0 M &975 G &5.9\\
\textbf{\cbname Swin-T}  &$\mathbf{53.6^{+3.1}}$ &$\mathbf{46.2^{+2.5}}$ &\textbf{113.8 M} &\textbf{836 G} &\textbf{6.5}\\
\bottomrule 
\end{tabular}
\label{tab:transform} 
\end{table}
}
\CheckRmv{
\begin{table}[t]
\centering
\tabFormat
\caption{\ttl{Generalization capability of CBNet over various special backbones.}} 
\vspace{-1mm}
\begin{tabular}{ccccc}
\toprule
\textbf{\ttl{Detector}}            & \textbf{\ttl{Backbone}}         & \textbf{\ttl{mAP}}  & \textbf{\ttl{Params}}   & \textbf{\ttl{FLOPs}}    \\ \midrule
\multirow{3}{*}{\ttl{YOLOV3}}      & \ttl{MobileNetV2} & \ttl{22.2} &\ttl{3.74 M} & \ttl{1.69 G} \\
& \textbf{\ttl{CB-MobileNetV2}}   & \textbf{\ttl{25.4}} & \textbf{\ttl{5.20   M}} & \textbf{\ttl{2.27   G}} \\
                             & \ttl{MobileNetV2   (1.4x)}      & \ttl{24.4}          & \ttl{5.04   M}          & \ttl{2.21   G}          \\\midrule
\multirow{3}{*}{\ttl{Faster-RCNN}} &  \ttl{HRNetv2p\_w32} &\ttl{40.2}& \ttl{47.3 M} &\ttl{285 G}\\
& \textbf{\ttl{CB-HRNetv2p\_w32}} & \textbf{\ttl{42.6}} & \textbf{\ttl{76.8   M}} & \textbf{\ttl{449   G}}  \\
                             & \ttl{HRNetv2p\_w48}              & \ttl{42.0}          & \ttl{83.4   M}          & \ttl{460   G}           \\\midrule
\multirow{6}{*}{\ttl{RetinaNet}} & \ttl{PVT-Small}& \ttl{40.4}&\ttl{34.2 M} & \ttl{214 G}\\
& \textbf{\ttl{CB-PVT-Small}}     & \textbf{\ttl{43.4}} & \textbf{\ttl{58.9   M}} & \textbf{\ttl{278   G}}  \\
                             & \ttl{PVT-Large}                 & \ttl{42.6}          & \ttl{71.1   M}          & \ttl{309   G}           \\ \cmidrule{2-5}
                             & \ttl{PVTv2-B2} & \ttl{44.6}& \ttl{35.1 M}& \ttl{218 G}	\\
                             & \textbf{\ttl{CB-PVTv2-B2}}      & \textbf{\ttl{47.7}} & \textbf{\ttl{60.7   M}} & \textbf{\ttl{287   G}}  \\
                             & \ttl{PVTv2-B5}                  & \ttl{46.1}          & \ttl{91.7   M}          & \ttl{335   G}           \\
\bottomrule
\end{tabular}
\label{tab:hrnet}
\end{table}
}

{
\subsubsection{Generality for special backbones} \label{sec:hrnet} To further show the generality of CBNet for various backbones, we conduct experiments on CBNet equipped with different backbones including MobileNetV2~\cite{SandlerHZZC18}, HRNet \cite{DBLP:journals/pami/00010CJDZ0MTW0X21}, PVT \cite{WangX0FSLL0021}, and PVTv2~\cite{wang2022pvt}.
For a fair comparison, we choose the publicly available pre-trained backbones and all experiment settings (\eg, choice of detectors, training, and inference details) are following their settings in MMDetection~\cite{chen2019mmdetection}. 
Results are shown in \tabref{tab:hrnet}.
For Mobile settings with YOLOV3~\cite{redmon2018yolov3}, our CB-MobileNetV2 improves MobileNetV2 by 3.1\% AP and is 1\% AP higher than MobileNetV2(1.4x) with comparable model complexity. 
For backbones with high-resolution representations, our CB-HRNetv2p$\_$w32 improves HRNetv2p$\_$w32 by 2.4\% AP and is 0.6\% AP higher than HRNetv2p$\_$w48 with less model complexity.
For global transformer backbones, we choose RetinaNet as detector follow the original paper~\cite{WangX0FSLL0021, wang2022pvt}. Our CB-PVT-Small improves PVT-Small by 3\% AP and is 0.8\% AP higher than PVT-Large with only 83\% number of parameters. Furthermore, our CB-PVTv2-B2 improves PVTv2-B2 by 3.1\% AP and is 1.6\% AP higher than PVTv2-B5 with only 66\% number of parameters. 
The results show that our CBNet improves a wide variety of backbones and achieves better accuracy under comparable or less parameters and FLOPs, which verify the effectiveness and efficiency of CBNet.
}

\subsubsection{Model Adaptability for Mainstream Detectors}
\label{sec:exp_detectors}
We evaluate the adaptability of CBNet by plugging it into mainstream detectors such as RetinaNet, ATSS, Faster R-CNN, Mask R-CNN, and Cascade R-CNN. These methods present a variety of detector head designs (\textit{e.g.,} two-stage vs. one-stage, anchor-based vs. anchor-free). As shown in \tabref{tab: detectors}, our CBNet significantly boosts all popular object detectors by over 3\% AP. The instance segmentation accuracy of Mask R-CNN is also improved by 2.9\%  AP. These results demonstrate the robust adaptability of CBNet to various head designs of detectors.

\CheckRmv{
\begin{table}[t!]
\tabFormat
\tabcolsep 0.05 in{\scriptsize{}}
\caption{Comparison of ResNet50 and our \cbname ResNet50.
\cbname ResNet50 significantly boosts all popular object detectors by 3.0\% $\sim$ 3.8\% bbox AP and 2.9\% mask AP. 'R50' is short for 'ResNet50'.
} 
\vspace{-2mm}
\begin{tabular}{c|ccc|ccc} 
\toprule
\multirow{2}{*}{\textbf{Detector}} & \multicolumn{2}{c}{$\mathbf{AP^{box}}$} & \multirow{2}{*}{$\Delta\mathbf{AP}$} & \multicolumn{2}{c}{$\mathbf{AP^{mask}}$} & \multirow{2}{*}{$\Delta\mathbf{AP}$} \\ 
   & R50  & CB-R50  & & R50  & CB-R50   & \\ \hline
RetinaNet  & 33.2 & \textbf{36.2} & +3.0 & -   & -  & -   \\
ATSS   & 36.9 & \textbf{39.9} & +3.0 & -   & -  & -   \\
Faster   R-CNN & 34.6 & \ttl{\textbf{38.0}} & +3.4   & -   & -  & -   \\
Mask   R-CNN   & 35.2 & \textbf{39.0} & +3.8   & 31.8 & \textbf{34.7}  & +2.9 \\
Cascade R-CNN  & 38.2 & \textbf{41.2} & +3.0 & -   & -  & -  \\
\bottomrule 
\end{tabular}
\label{tab: detectors} 
\end{table}
}

\CheckRmv{
\begin{table}[t]
\centering
\tabFormat
\caption{\ttl{Comparison of CBNet with YOLOX and Joint-DetNAS.}} 
\vspace{-1mm}
\begin{tabular}{ccccc}
\toprule
\textbf{\ttl{Backbone}}  &\textbf{\ttl{Method}}                 & \textbf{\ttl{mAP}}  & \textbf{\ttl{Params}}   & \textbf{\ttl{FLOPs}}    \\ \midrule

\ttl{CSPNet-L} &\multirow{3}{*}{\ttl{YOLOX}} & \ttl{49.4} & \ttl{54.2 M} & \ttl{78 G}\\ 
\ttl{CSPNet-X}      &             & \ttl{50.9}          & \ttl{99.1   M}          & \ttl{141   G}         \\
\textbf{\ttl{CB-CSPNet-L}}    &   & \textbf{\ttl{52.0}} & \textbf{\ttl{83.8   M}} & \textbf{\ttl{118   G}}  \\
\midrule
\textbf{\ttl{Searched X101-FPN}}    &\ttl{Joint-DetNAS}   & \textbf{\ttl{45.7}} & \textbf{\ttl{-}} & \textbf{\ttl{266 G}}  \\

\bottomrule
\end{tabular}
\label{tab:yolox}
\end{table}
}

\begin{table}[]
\centering
\caption{\ttl{Comparison between CBNet and DetecoRS based on Faster-RCNN with $1\times$ training. CBNet achieves on par or better accuracy-efficiency trade-offs.}} \vspace{-2mm}
\begin{tabular}{ccccc}
\toprule
\textbf{\ttl{Backbone}}                 & \textbf{\ttl{Method}} & \textbf{\ttl{mAP}}  & \textbf{\ttl{Params}}    & \textbf{\ttl{FLOPs}}   \\ \midrule
\multirow{2}{*}{\ttl{ResNet101}}        & \textbf{\ttl{CBNet}}  & \textbf{\ttl{42.7}} & \textbf{\ttl{107.4   M}} & \textbf{\ttl{436   G}} \\
                                  & \ttl{DetectoRS}       & \ttl{42.8}          & \ttl{104.0   M}          & \ttl{512   G}          \\\midrule
\multirow{2}{*}{\ttl{ResNeXt101-32x4d}} & \textbf{\ttl{CBNet}}  & \textbf{\ttl{44.2}} & \textbf{\ttl{106.7   M}} & \textbf{\ttl{444   G}} \\
                                  & \ttl{DetectoRS}       & \ttl{44.0}            & \ttl{103.5   M}          & \ttl{519   G}          \\\midrule
\multirow{2}{*}{\ttl{Res2Net101}}       & \textbf{\ttl{CBNet}}  & \textbf{\ttl{45.8}} & \textbf{\ttl{108.7   M}} & \textbf{\ttl{452   G}} \\
                                  & \ttl{DetectoRS}       & \ttl{45.7}          & \ttl{105.4   M}          & \ttl{533   G}          \\\midrule
\multirow{2}{*}{\ttl{Swin-Tiny}}        & \textbf{\ttl{CBNet}}  & \textbf{\ttl{46.7}} & \textbf{\ttl{74.1   M}}  & \textbf{\ttl{307   G}} \\
                                  & \ttl{DetectoRS}       & \ttl{45.9}          & \ttl{73.4   M}           & \ttl{366   G}         \\ \bottomrule
                                 
\end{tabular}
\label{tab:rs}
\end{table}
{
\subsection{Comparison with Relevant Works.}
There are several relevant detectors, such as DetectoRS \cite{QiaoCY21} that composites both backbone and FPN and Joint-DetNAS \cite{YaoPXZLZ21} searches for the model scaling strategy. We conduct comparisons between CBNet and these two methods.

Joint-DetNAS \cite{YaoPXZLZ21} integrates neural architecture search (NAS), pruning and knowledge distillation for optimizing detectors. 
Similarly, our CBNet also uses pruning strategy but focus more on scaling backbones using composite strategy. Thanks to the strong generalization ability, our CBNet can boosts the performance of advanced high-performance detectors (\eg, YOLOX~\cite{yolox2021}). As shown in Table~\ref{tab:yolox}, our CB-CSPNet-L improves CSPNet-L~\cite{yolox2021} by 2.6\% AP and is 1.1\% AP higher than CSPNet-X with only 85\% number of parameters. 
We further compare our CBNet using an existing hand-designed detector (\ie, YOLOX) with Joint-DetNAS which uses an advanced knowledge distillation training strategy. Our CBNet achieves 52\% AP with 118 GFLOPs, superior to that of Joint-DetNAS (X101-FPN based) at 45.7\% AP with 266 GFLOPs. Note that it is hard to have a fair comparison because our CBNet focuses on the architecture design of the backbone while Joint-DetNAS focuses on the joint optimization of the architecture and training for the entire detector.

DetectoRS \cite{QiaoCY21} conducts a similar design as CBNet while DetectoRS composites both backbone and FPN. We compare CBNetV2 and DetectoRS with different backbones on Faster R-CNN in \tabref{tab:rs}.  Under the same training strategy \ttl{of DetectoRS with $1333\times 800$ as input size}, CBNet achieves comparable or higher AP with fewer FLOPs. Specifically, with advanced backbone Swin-Tiny, our CBNet outperforms DetectoRS by 0.8\% AP with only 84\% FLOPs.
}

\subsection{Compatibility of CBNet}
\label{sec:compatible}

\subsubsection{Compatibility  with Deformable Convolution}
Deformable convolution \cite{DaiQXLZHW17} enhances the transformation modeling capability of CNNs and is widely used for accurate object detectors  (\textit{e.g.,} simply adding DCN improves Faster R-CNN ResNet50 from 34.6\% to 37.4\% AP). To show the compatibility of CBNet architecture with deformable convolution, we perform experiments on ResNet and ResNeXt equipped with Faster R-CNN. As shown in \tabref{tab:dcn}, DCN is still effective on CBNet with 2.3\% AP$\sim$2.7\% AP improvement. This improvement is greater than the 2.0\% AP and 1.3\% AP increments on ResNet152 and ResNeXt101-64x4d. On the other hand, \cbname ResNet50-DCN increases the AP of ResNet50-DCN and the deeper ResNet152-DCN by 3.0\% and 0.6\%, respectively. In addition, \cbname ResNet50-32x4d-DCN increases the AP of ResNet50-32x4d-DCN and the deeper and wider ResNeXt101-64x4d-DCN by 3.7\% and 1.3\%, respectively. 
The results show that the effects of CBNet and deformable convolution can be superimposed without conflicting with each other. 
\CheckRmv{
\begin{table}[t!]
\tabFormat
\setlength{\tabcolsep}{2.3mm}
\caption{Experimental results on the compatibility of CBNet and deformable convolution. CBNet and deformable convolution can be superimposed on each other without conflict. }
\vspace{-2mm}
\begin{tabular}{ccccc} 
\toprule
\multirow{2}{*}{\textbf{Backbone}} & \multicolumn{2}{c}{$\mathbf{AP^{box}}$} & \multirow{2}{*}{$\Delta\mathbf{AP}$} \\
& w/o DCN & w/ DCN &  \\\hline
ResNet50 & 34.6 & 37.4 & +2.8 \\
ResNet152  & 37.8 & 39.8 & +2.0  \\
\textbf{\cbname ResNet50} & \ttl{\textbf{38.0}} & \textbf{40.4} & \textbf{+2.4} \\\hline
ResNeXt50-32x4d  & 36.3 & 38.6 & +2.3 \\
ResNeXt101-64x4  & 38.7 & 41.0 & +2.3 \\
\textbf{\cbname ResNeXt50-32x4} & \textbf{39.8} & \textbf{42.5} & \textbf{+2.7} \\
\bottomrule 
\end{tabular}
\label{tab:dcn} 
\end{table}
}
\CheckRmv{
\begin{table}[t!]
\tabFormat
\setlength{\tabcolsep}{2.3mm}
\caption{Compatibility of CBNet and Model Ensemble Method. They can be superimposed on each other without conflict. 'R50', 'X50', 'R2-50', 'R101', 'X101', 'R2-101' are short for Faster R-CNN equipped with ResNet50, ResNeXt50-32x4d, Res2Net50, ResNet101, ResNeXt101-64x4, Res2Net101 respectively. 
'Single' denotes single model best AP.}
\vspace{-2mm}
\begin{tabular}{c|cc|c}
\toprule
\multirow{2}{*}{\textbf{Ensemble   Models}}      & \multicolumn{2}{c|}{$\mathbf{AP^{box}}$}       & \multirow{2}{*}{$\Delta\mathbf{AP}$} \\
& \textbf{Single}  & \textbf{Ensem.}      &                              \\
\hline
R50(34.6) \& X50(36.3)                    & 36.3          & 37.3          & 1.0                          \\
R101(36.3) \& X101(38.7)                  & 38.7          & 39.7          & 1.0                          \\
\textbf{\cbname R50(38.0) \&   \cbname X50(39.8)}   & \textbf{39.8} & \textbf{40.8} & \textbf{1.0}                 \\\hline
R50(34.6) \& R2-50(37.7)                  & 37.7          & 38.8          & 1.1                          \\
R101(36.3) \& \ttl{R2-101(40.1)}                  & 40.1          & 40.9          & 0.8                          \\
\textbf{\cbname R50(38.0) \&   \cbname R2-50(41.2)} & \textbf{41.2} & \textbf{42.2} & \textbf{1.0}                 \\\hline
X50(36.3) \& R2-50(37.7)                  & 37.7          & 39.4          & 1.7                          \\
X101(38.7) \& R2-101(40.1)                & 40.1          & 41.9          & 1.8                          \\
\textbf{\cbname X50(39.8) \&   \cbname R2-50(41.2)} & \textbf{41.2} & \textbf{42.9} & \textbf{1.7}                \\\hline
\end{tabular}
\label{tab:ensem} 
\end{table}

}

\subsubsection{Compatibility with Model Ensemble}
\label{sec:ensem}
The model ensemble improves the prediction performance of a single model by training multiple different models and combining their prediction results through post-processing \cite{nips/KroghV94, widm/SagiR18}. Probabilistic Ranking Aware Ensemble (PRAE) \cite{DBLP:journals/corr/abs-2105-03139}  refines the confidence of bounding boxes from different detectors and outperforms other ensemble learning methods for object detection by significant margins (\textit{e.g.,} assembling Faster R-CNN ResNet50 and Faster R-CNN ReNeXt50 improves the single model best AP from 36.3\% to 37.3\%), \ttl{Note that assembling two same detectors (i.e., two Faster R-CNN ResNeXt50) does not improve the performance (same as the single detector 36.3\% AP).} To show the compatibility of our CBNet architecture with the model ensemble method PRAE, we perform experiments on traditional backbones (\ie, ResNet, ResNeXt, Res2Net) and their Composite Backbones equipped with Faster R-CNN. As shown in \tabref{tab:ensem}, PRAE is still effective for assembling detectors with CBNet, with 0.8\% $\sim$ 1.7\% AP improvement, which is consistent with the case of assembling detectors with traditional backbones. \ttl{In addition, CBNet is more effective than the model ensembling method PRAE, e.g., Faster R-CNN CB-R50 achieves 38.0\% AP, superior to the 37.3\% AP of assembling Faster R-CNN ResNet50 and Faster R-CNN ReNeXt50.} The results show that the effects of CBNet and model ensemble can be superimposed without conflicting with each other, suggesting that the detector equipped with CBNet should be considered as a single detector/model despite having multiple identical backbones compositions.

\subsection{Ablation Studies}
\label{sec:ablation}
We ablate various design choices for our proposed CBNet. For simplicity, all accuracy results here are on the COCO validation set with  $800\times 500$ input size if not specified.

\subsubsection{Effectiveness of Different Composite Strategies}
\label{exp:composite_style}
We conduct experiments to compare the proposed composite strategies in Fig.~\ref{fig:comp_style}, including SLC, AHLC, ALLC, DHLC and FCC. All these experiments are conducted based on the Faster R-CNN \cbname ResNet50 architecture. Results are shown in \tabref{tab:compstyle}.

\textbf{SLC} gets a slightly improves accuracy of the single-backbone baseline (35\% vs. 34.6\% AP). 
The features extracted by the same stage of both backbones are similar, and thus SLC can only learn slightly more semantic information than a single backbone does. 

\textbf{AHLC} raises the baseline by 1.4\% AP, which verifies our motivation in \secref{sec:ahlc}, \ie, the semantic information higher-level features of the former backbone enhances the representation ability of the latter backbone.

\textbf{ALLC} degrades the performance of the baseline by 2.2\% AP.  We infer that directly adding the lower-level  
features of the assisting backbone to the higher-level  
ones of the lead backbone impair the representation ability of the latter. 

\textbf{DHLC} improves the performance of the baseline by a large margin  (from 34.6\% AP to 37.3\% AP by 2.7\% AP). More composite connections of the high-low cases enrich the representation ability of features to some extent.

\textbf{FCC} achieves the best performance of 37.4\% AP while being 7\% slower than DHLC (19.9 vs. 21.4 FPS).

\CheckRmv{
\begin{table}[t!]
\tabFormat
\setlength{\tabcolsep}{2mm}
\caption{Comparison between different composite strategies. Obviously, DHLC acheives the best FLOPs-accuracy and Params-accuracy trade-offs.
} 
\vspace{-2mm}
\begin{tabular}{c  c  c c c} 
\toprule
\textbf{Composite Strategy} & $\mathbf{AP^{box}}$ &  \textbf{Params} &\textbf{FLOPs} &\textbf{\ttl{FPS}}\\ 
\hline	
- & 34.6  & 41.5 M &90 G &\ttl{30.0}\\
\hline
SLC & 35.0 &64.8 M&123 G &\ttl{22.6}\\
AHLC & 36.0  &67.6 M &126 G &\ttl{22.4}\\
ALLC & 32.4 &67.6 M &133 G &\ttl{22.5}\\
\textbf{DHLC}	&\textbf{37.3}&69.7 M &127 G &\ttl{21.4}\\
\textbf{FCC}	& \textbf{37.4} &72.0 M &145 G &\ttl{19.9}\\
\bottomrule 
\end{tabular}
\label{tab:compstyle} 
\end{table}
}

\CheckRmv{
\begin{table}[t!]
\tabFormat
\setlength{\tabcolsep}{1mm}
\caption{\ttl{Comparison between AHLC and DHLC on different backbones.}
} 
\vspace{-2mm}
\begin{tabular}{c c c  c c} 
\toprule
\textbf{\ttl{Backbone}} &\textbf{\ttl{Composite}} & \ttl{$\mathbf{AP^{box}}$} &  \textbf{\ttl{Params}} &\textbf{\ttl{FLOPs}}\\ 
\hline	
\multirow{2}{*}{\textbf{\ttl{\cbname ResNet101}}}
&\ttl{AHLC} &\ttl{37.9}	&\ttl{105.6 M} &\ttl{186 G}	\\
&\textbf{\ttl{DHLC}}	&\textbf{\ttl{38.7}}	&\ttl{107.6 M} &\ttl{188 G}	\\\hline
\multirow{2}{*}{\textbf{\ttl{\cbname ResNeXt101}}}
&\ttl{AHLC} &\ttl{39.9}	&\ttl{183.0 M} &\ttl{312 G}	\\
&\textbf{\ttl{DHLC}}	&\textbf{\ttl{41.0}}	&\ttl{185.1 M} &\ttl{313 G}	\\
\bottomrule 
\end{tabular}
\label{tab:ahlc_dhlc} 
\end{table}
}

In summary, FCC and DHLC achieve the two best results. 
Considering the computational simplicity, we recommend using DHLC for CBNet. 
All the above composite strategies have a similar amount of parameters, but the accuracy varies greatly.
The results prove that simply increasing the number of parameters or adding a backbone network does not  guarantee a better result while the composite connection plays a crucial part. These results show that the suggested DHLC composite strategy is effective and nontrivial.

{Note that there is a minor performance gap between the original CBNetV1 in~\cite{LiuWWLZTL20} and this paper for DHLC and AHLC. The reason is that CBNetV1 and this paper are performed under different deep learning platforms (CAFFE vs. PyTorch), which use different model initialization strategies and result in different model performances.  We compare DHLC and AHLC on different backbones in \tabref{tab:ahlc_dhlc}. DHLC outperforms AHLC by 0.8\% AP and 1.1\% AP on ResNet101 and ResNeXt101-64x4d, respectively, showing the generality of DHLC for different backbones.

We conduct a grid search by proxy task to search for better composite strategies. To reduce the search cost, we simplify the search space by only searching the connections including $x_3, x_4, x_5$ stages in composite backbones, and design a proxy task with 1/5 of the COCO training set with input size set to $800\times 500$. In this way, we only need to train $(2^3 )^3=512$ detectors for 205 GPU days. The best-searched strategy is a simplified DHLC($s_3$) without the connection between $x_4$ of the former backbone to the input of $x_3$ of the latter one. The searched strategy achieves 37.3\% AP with 69.1 M, 126 GFLOPs, and performs un-par with our designed DHLC (37.3\% AP with 69.7 M, 127 GFLOPs), further validating the necessity of high-to-low connections in our handcraft design.
}

\subsubsection{Weights for Auxiliary Supervision}
\label{exp:supervise}
Experimental results related to weighting the auxiliary supervision are presented in \tabref{tab:supervisew}. For simplicity, we perform DHLC composite strategy on CBNet. The first setting is the Faster R-CNN \cbname ResNet50 baseline and the second is the \cbname ResNet50-K3 ($K=3$ in CBNet) baseline, where the  $\lambda$ for assisting backbone in \eqref{eq:supervision} is set to zero. For the case $K=2$, the baseline can be improved by 0.8\% AP by setting $\lambda_1$ to 0.5. For the case $K=3$, the baseline can be improved by 1.8\% AP by setting $\{\lambda_1, \lambda_2\}$ to $\{0.5, 1.0\}$. The experimental results verify that the auxiliary supervision forms an effective training strategy that improves the performance of CBNet.
\CheckRmv{
\begin{table}[t!]
\tabFormat
\setlength{\tabcolsep}{2.3mm}
\caption{Ablation study of loss weights for assistant supervision.
}
\vspace{-2mm}
\label{tab:supervisew}
\begin{tabular}{cccc}
\toprule
\textbf{Backbone} & $\mathbf{\lambda_1}$ & $\mathbf{\lambda_2}$ & $\mathbf{AP^{bbox}}$ \\ \hline
\multirow{5}{*}{\textbf{\cbname ResNet50}} & 0  & -  & 37.3 \\ 
& 0.125  & -  & 37.6 \\ 
& 0.25 & -  & 37.8 \\ 
& \textbf{0.5}  & \textbf{-} & \textbf{38.1} \\ 
& 1.0 & -  & 37.9 \\ \hline
\multirow{5}{*}{\textbf{\cbname ResNet50-K3}} & 0  & 0  & 37.4 \\ 
& 0.25 & 0.25 & 37.9 \\ 
& 0.25 & 0.5 & 38.9 \\ 
& 0.5 & 0.5 & 38.6 \\ 
& \textbf{0.5}  & \textbf{1.0}  & \textbf{39.2} \\
\bottomrule
\end{tabular}
\end{table}
}
\CheckRmv{
\begin{table}[t!]
\tabFormat
\setlength{\tabcolsep}{1mm}
\caption{Comparison between CBNetV1  \cite{LiuWWLZTL20} and CBNetV2. CBNetV2 is superior to CBNetV1 in terms of accuracy and complexity.}
\vspace{-2mm}
\label{tab:total}
\begin{tabular}{cccccccc}
\toprule
\textbf{Backbone} &\textbf{DHLC} & \textbf{Sup.} & \textbf{Prun.} & $\mathbf{AP^{box}}$ &\textbf{Params} &\textbf{FLOPs} &\textbf{\ttl{FPS}}\\ \hline
ResNet50 & & & & 34.6 &41.5 M &90 G &\ttl{30.0}\\\hline 
CBNetV1 & & & &36.0 &67.6 M & 126 G &\ttl{22.4}\\ 
\ttl{CBNetV1} & & &\ttl{\ding{51}} &\ttl{35.6} &\ttl{66.2 M} & \ttl{111 G} &\ttl{26.6}\\ 
& & \ding{51} & & 36.9  &67.6 M & 126 G &\ttl{22.4}\\ 
\hline
&\ding{51} & & & 37.3  &69.7 M & 127 G &\ttl{21.4}\\ 
 &\textbf{\ding{51}} & \textbf{\ding{51}} & & \textbf{38.1}  &69.7 M & 127 G &\ttl{21.4}\\ 
CBNetV2 &  \textbf{\ding{51}} & \textbf{\ding{51}} & \textbf{\ding{51}} & \textbf{38.0}  &69.4 M & \textbf{121 G} &\ttl{23.3}\\ 
\bottomrule
\end{tabular}
\end{table}
}

\subsubsection{Efficiency of Pruning Strategy}
\label{sec:prune}

As shown in \figref{fig:dele}, with the pruning strategy, our \cbname ResNet50 family and \cbname ResNet50-K3 family achieve better FLOPs-accuracy trade-offs than ResNet family. This also illustrates the efficiency of our pruning strategy. In particular, the number of FLOPs in $s_3$ is reduced by 10\% compared to $s_4$, but the accuracy is decreased by only 0.1\%. This is because the weights of the pruned stage are fixed during the detector training \cite{mmdetection} so pruning this stage does not sacrifice detection accuracy.
Hence, when speed and memory cost need to be prioritized, we suggest pruning the fixed stages in $2, 3, ..., K$-th backbones in CBNet.

\subsubsection{Number of Backbones in CBNet}
To further explore the ability to construct high-performance detectors of CBNet, we evaluate the efficiency of our CBNet by controlling the number of backbones. As shown in \figref{fig:num}, we vary the number of backbones (\textit{e.g.,} $K$ = 1,2,3,4,5) and compare their accuracy and efficiency (GFLOPs) with the ResNet family. Note that the accuracy continues to increase as the complexity of the model increases.
Compared with ResNet152, our method obtains higher accuracy at $K$=2 while computation cost is lower. Meanwhile, the accuracy can be further improved for $K$=3,4,5. CBNet provides an effective and efficient alternative to improve the model performance rather than simply increasing the depth or width of the backbone.

\subsubsection{Comparison of CBNetV1 and CBNetV2}
\label{exp:component}
To fairly compare CBNetV1\cite{LiuWWLZTL20} and CBNetV2, we progressively apply the DHLC composite strategy, auxiliary supervision, and pruning strategy to CBNetV1, where AHLC is the default composite strategy in \tabref{tab:total}.
{As in the 1st and 2nd rows of \tabref{tab:total}}, the composite backbone structure CBNetV1  \cite{LiuWWLZTL20} improves the Faster R-CNN ResNet50 baseline by 1.4\% AP.  \ttl{As in the 2nd and 3rd row, the accelerated version of CBNetV1 ($s_2$ pruning version in \figref{fig:simpcb}) improves the inference speed from 22.4 FPS to 26.6 FPS while decreasing the accuracy by 0.4\% AP.}
{As in the 2nd and 4th rows,} the auxiliary supervision brings a 0.9\% AP increment to CBNetV1, thanks to the better training strategy that improves the representative ability of the lead backbone. Note that the auxiliary supervision does not introduce extra parameters during the inference phase.
{As in the 2nd and 5th rows,} DHLC composite strategy improves the detection performance of CBNetV1 by 1.3\% AP with higher model complexity. The results confirm that DHLC enables a larger receptive field, with features at each level obtaining rich semantic information from all higher-level features. 
{As in the 1st and 6th rows,} when combining the DHLC and the auxiliary supervision, there is a significant improvement of 2.1\% AP over the baseline. \ttl{As in the 2nd and last row,} when we perform our default pruning strategy ($s_3$ version in \figref{fig:simpcb}), 
{CBNetV2 is faster (23.3 vs. 22.4 FPS) and much more accurate (38.0\% vs. 36.0\% AP) than CBNetV1 \cite{LiuWWLZTL20}. DHLC slows down the detector, while the pruning strategy effectively speeds up the inference speed of CBNetV2.}

\begin{figure}
     \centering
     \begin{subfigure}[b]{0.24\textwidth}
         \centering
         \includegraphics[width=\textwidth]{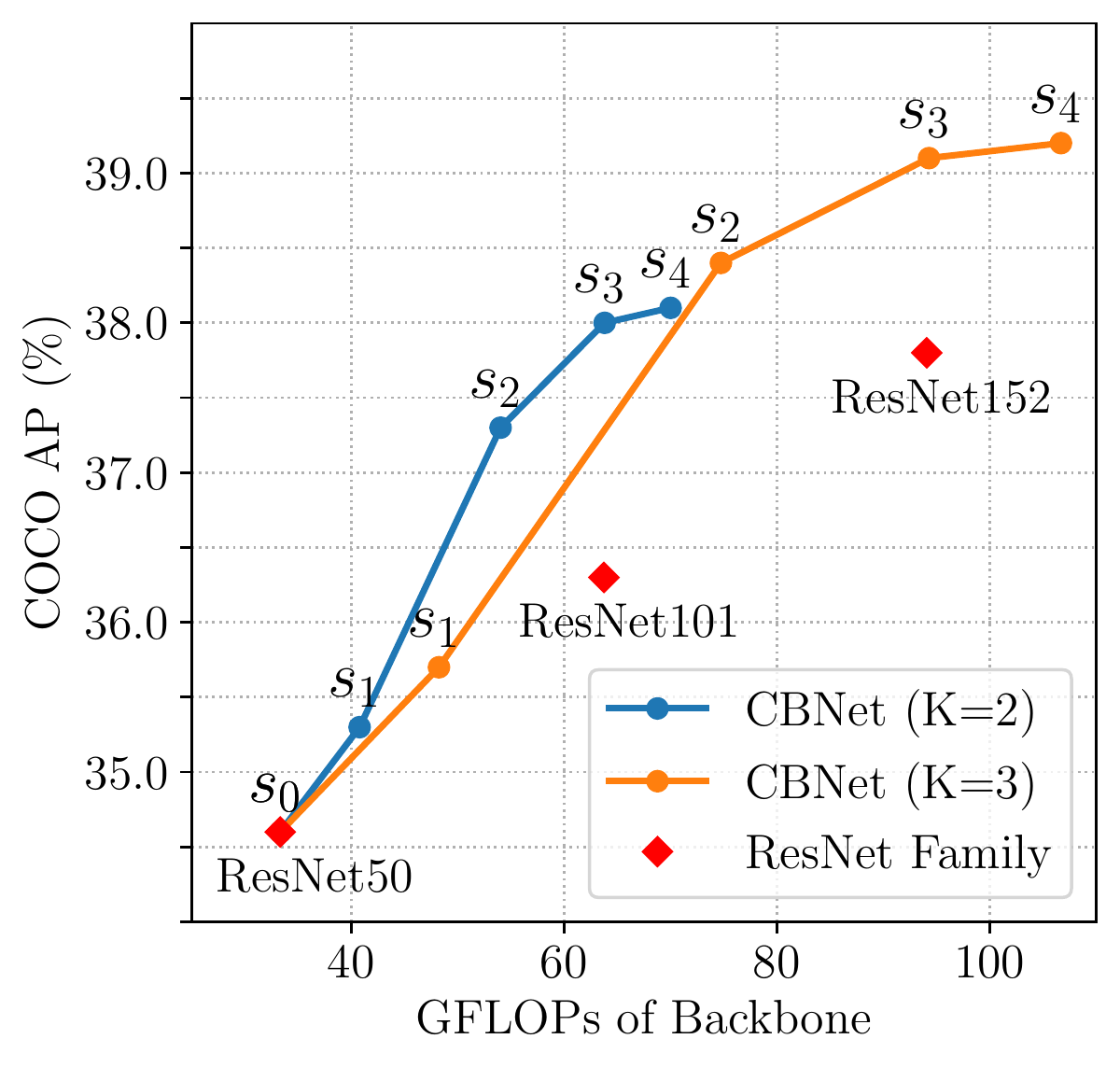}
         \vspace{-6mm}
         \caption{Pruning strategies}
         \label{fig:dele}
     \end{subfigure}
     \hfill
     \begin{subfigure}[b]{0.24\textwidth}
         \centering
         \includegraphics[width=\textwidth]{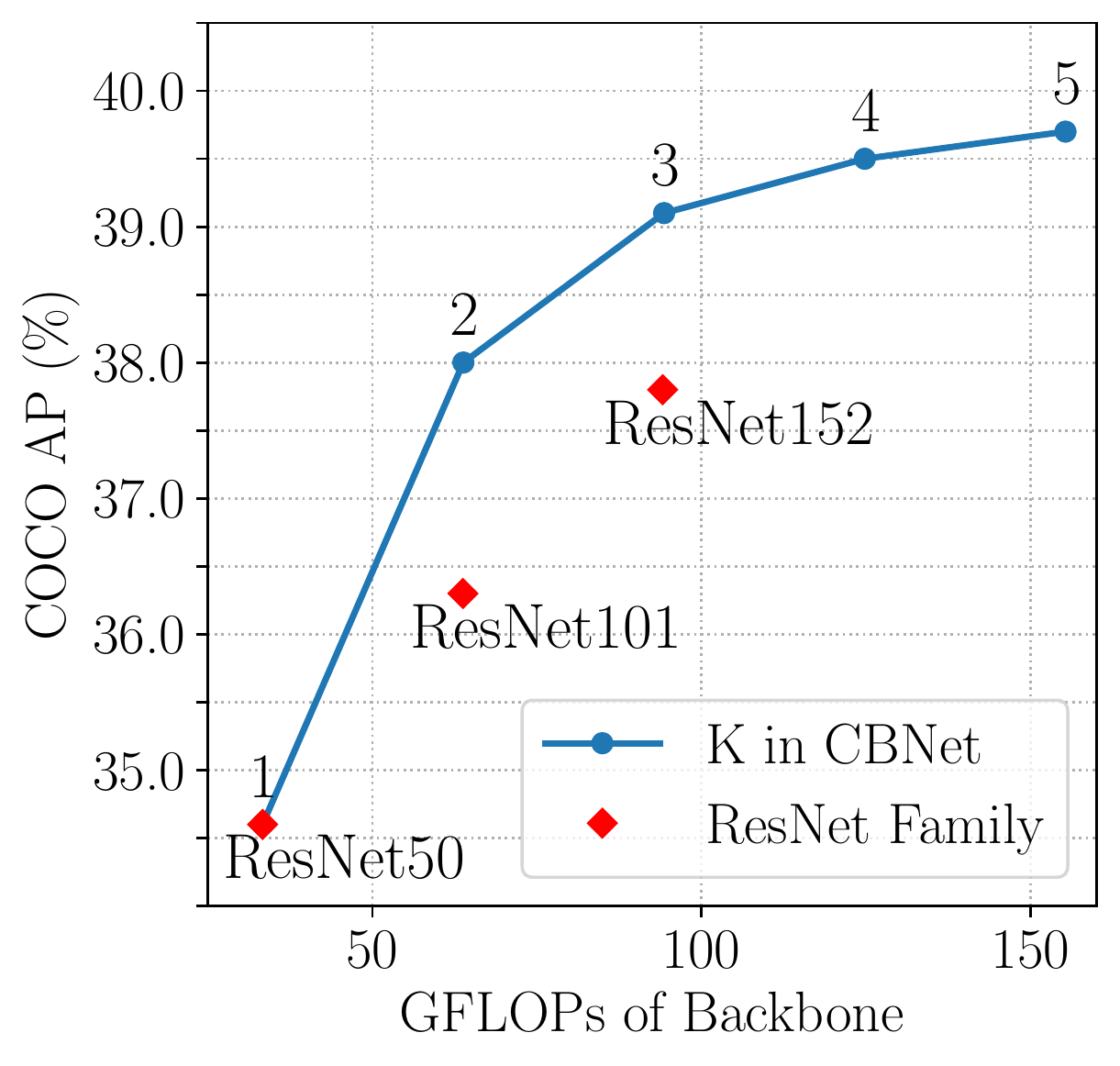}
         \vspace{-6mm}
         \caption{\#Backbones ($K$)}
         \label{fig:num}
     \end{subfigure}
    \caption{Performance comparison of CBNet with different numbers of composite backbones ($K$) and pruning strategies. }
\end{figure}

\subsubsection{Importance of Identical Backbones for CBNet}
\label{sec:identical}
To verify the necessity of identical backbones in CBNet, we explore the diversity backbones by 
\ttl{compositing  ResNet50, ResNet101, Res2Net50, and Res2Net101.
Note that no pruning is conducted for compositing diverse backbones and backbones from different families do not share the stem layer (Conv1 in \figref{fig:RCNN}).} As shown in \tabref{tab:50-c-101},
\ttlv{for backbones belonging to the same family, compositing identical backbones outperforms compositing diverse ones. For example, \cbname ResNet50 achieves higher AP with fewer parameters than both ResNet50-C-ResNet101 and ResNet101-C-ResNet50. Similarly, \cbname Res2Net50 gains higher or comparable AP with  fewer parameters than both Res2Net50-C-Res2Net101 and Res2Net101-C-Res2Net50.  
For backbones from different families, the observation still holds. For example, CB-Res2Net50 achieves better performance than ResNet50-C-Res2Net101, Res2Net101-C-ResNet50, ResNet101-C-Res2Net50, and Res2Net50-C-ResNet101. 
}
These experimental results indicate that increasing the diversity of composite models \ttl{is not the most efficient way for CBNet.}
\ttlv{We believe the reason is that using different backbones needs different optimization strategies, which usually output very different learned features and are difficult for joint training. CBNet intends to learn similar features for each grouped backbone, and the stronger the former backbones are, the more representative features the lead backbone outputs. Our experiments show that such a joint-training strategy works best for identical backbone grouping. }
This validates the necessity of the identical backbones in CBNet and further distinguishes our approach from ensemble methods where diversity is a key character. 
}

\CheckRmv{\section{Conclusion}
\label{sect:conclu}
In this paper, we propose a novel and flexible backbone framework, called \emph{Composite Backbone Network } (CBNet), to improve the performance of cutting-edge object detectors. 
\CheckRmv{
\begin{table}[t!]
\tabFormat
\setlength{\tabcolsep}{2.3mm}
\caption{Importance of identical backbones for CBNet. Compositing two identical ResNet50 achieves better performance than compositing ResNet50 and ResNet101. }
\vspace{-2mm}
\begin{tabular}{cccc} 
\toprule
\textbf{Backbone}   & $\mathbf{AP^{box}}$ &  \textbf{Params} &\textbf{FLOPs} \\ 
\hline
ResNet50-C-ResNet101  & 37.8  & 88.7 M &157 G\\
ResNet101-C-ResNet50  & 37.8  & 88.7 M &157 G \\
\textbf{\cbname ResNet50}  &\textbf{38.0} &\textbf{69.4 M} &\textbf{121 G}\\\hline
\ttlv{Res2Net50-C-Res2Net101} & \ttlv{41.1}  & \ttlv{89.5 M} &\ttlv{163 G} \\ 
\ttlv{Res2Net101-C-Res2Net50} & \ttlv{41.4}  & \ttlv{89.5 M} &\ttlv{163 G} \\ 
\textbf{\cbname Res2Net50}  &\textbf{41.2} &\textbf{\ttlv{69.7 M}} &\textbf{125 G}\\
\ttlv{\textbf{\cbname Res2Net101}} & \ttlv{\textbf{43.0}}  & \ttlv{\textbf{108.7 M}} &\ttlv{\textbf{188 G}} \\\hline
\ttlv{ResNet50-C-Res2Net101} & \ttlv{40.0}  & \ttlv{89.3 M} &\ttlv{162 G} \\
\ttlv{Res2Net101-C-ResNet50} & \ttlv{41.2}  & \ttlv{89.3 M} &\ttlv{162 G} \\
\ttlv{ResNet101-C-Res2Net50} & \ttlv{39.5}  & \ttlv{88.8 M} &\ttlv{162 G} \\
\ttlv{Res2Net50-C-ResNet101} & \ttlv{40.6}  & \ttlv{88.8 M} &\ttlv{162 G} \\
\textbf{\cbname Res2Net50}  &\textbf{41.2} &\textbf{\ttlv{69.7 M}} &\textbf{125 G}\\
\bottomrule
\end{tabular}
\label{tab:50-c-101} 
\end{table}
}
CBNet consists of a series of backbones with the same network architecture in parallel, the Dense Higher-Level composition strategy, and the auxiliary supervision. Together they construct a robust representative backbone network that uses existing pre-trained backbones under the pre-training fine-tuning paradigm.
CBNet has strong generalization capabilities for different backbones and head designs of the detector architecture.
Extensive experimental results demonstrate that the proposed CBNet is compatible with various backbones networks, including CNN-based (ResNet, ResNeXt, Res2Net) and Transformer-based (Swin-Transformer) ones. At the same time, CBNet is more effective and efficient than simply increasing the depth and width of the network. Furthermore, CBNet can be flexibly plugged into most mainstream detectors, including one-stage (\textit{e.g.,} RetinaNet) and two-stage (Faster R-CNN, Mask R-CNN, Cascade R-CNN, and Cascade Mask R-CNN) detectors, as well as anchor-based (\textit{e.g.,} Faster R-CNN) and anchor-free-based (ATSS) ones. CBNet is compatible with feature enhancing networks (DCN and HRNet) and model ensemble methods. Specifically, the performances of the above detectors are increased by over 3\% AP. 
In particular, our \cbname Swin-L achieves a new record of 59.4\% box AP and 51.6\% mask AP on COCO \texttt{test-dev}, outperforming prior single-model single-scale results. With multi-scale testing, we achieve a new state-of-the-art result of 60.1\% box AP and 52.3\% mask AP without extra training data.
}
\ifCLASSOPTIONcompsoc
  \section*{Acknowledgments}
\else
  \section*{Acknowledgment}
\fi
This work was supported by National Natural Science Foundation of China under Grant 62176007. This work was also a research achievement of Key Laboratory of Science, Technology, and Standard in Press Industry (Key Laboratory of Intelligent Press Media Technology). 
We thank Dr. Han Hu, Prof. Ming-Ming Cheng and Shang-Hua Gao for the insightful discussions.

\ifCLASSOPTIONcaptionsoff
  \newpage
\fi



%
\bibliographystyle{IEEEtran}
\bibliography{egbib}

%

%
%
%




\end{document}